\documentclass[11pt]{article}
\pdfoutput=1
\usepackage[final]{acl}
\usepackage{subcaption}  
\usepackage{times}
\usepackage{latexsym}
\usepackage{url}
\usepackage{enumitem}
\usepackage[T1]{fontenc}
\usepackage[utf8]{inputenc}
\usepackage{microtype}
\usepackage{amsmath}
\usepackage{inconsolata}
\usepackage{booktabs}
\usepackage{graphicx}
\usepackage{float}
\usepackage{listings}
\usepackage{xcolor}
\usepackage{hyperref}
\title{LLMs Do Not See Age: Assessing Demographic Bias in Automated Systematic Review Synthesis}
\author{
  Favour Yahdii Aghaebe\thanks{School of Computer Science}\thanks{Healthy Lifespan Institute},
  Tanefa Apekey\thanks{Sheffield Centre for Health and Related Research},
  Elizabeth Williams\thanks{Department of Oncology and Metabolism}\footnotemark[2],
  Nafise Sadat Moosavi\footnotemark[1] 
  \\
  University of Sheffield, UK \\
  \texttt{\{fyaghaebe1, t.apekey, e.a.williams, n.s.moosavi\}@sheffield.ac.uk}
}
\begin{document}
\maketitle
\begin{abstract}
Clinical interventions often hinge on age: medications and procedures safe for adults may be harmful to children or ineffective for older adults. However, as language models are increasingly integrated into biomedical evidence synthesis workflows, it remains uncertain whether these systems preserve such crucial demographic distinctions. To address this gap, we evaluate how well state-of-the-art language models retain age-related information when generating abstractive summaries of biomedical studies.
We construct \textbf{DemogSummary}, a novel age-stratified dataset of systematic review primary studies, covering child, adult, and older adult populations.
We evaluate three prominent summarisation-capable LLMs, Qwen (open-source), Longformer (open-source) and GPT-4.1 Nano (proprietary), using both standard metrics and a newly proposed \textbf{Demographic Salience Score (DSS)}, which quantifies age-related entity retention and hallucination. 
Our results reveal systematic disparities across models and age groups: demographic fidelity is lowest for adult-focused summaries, and under-represented populations are more prone to hallucinations. These findings highlight the limitations of current LLMs in faithful and bias-free summarisation and point to the need for fairness-aware evaluation frameworks and summarisation pipelines in biomedical NLP.
\end{abstract}

\section{Introduction}
The use of large language models (LLMs) to enhance the efficiency of scientific research and clinical practice has become increasingly common. In particular, LLMs have shown promise in accelerating labour-intensive processes such as systematic reviews. These models are now being explored at various stages of the review pipeline, including abstract identification and screening \citep{doi:10.1073/pnas.2411962122}, offering the potential to reduce time and cost in evidence synthesis.

Despite these advances, concerns are emerging about the potential biases introduced by LLMs when applied to domains involving sensitive or underrepresented populations. Recent studies suggest that LLMs may exhibit demographic bias in their outputs \citep{kamruzzaman-etal-2024-investigating}. In the context of systematic reviews, which often inform high-stakes decisions in medicine and policy, such biases pose serious risks. The process by which LLMs synthesise and preserve critical demographic information, specifically with narrative synthesis, where findings from individual studies are summarised, remains largely unexamined. 

This presents a critical gap: while LLMs are increasingly used in summarising biomedical literature, little is known about their ability to retain demographic and population information, particularly age-related descriptors which are integral aspects of the systematic review. Misrepresentation or omission of such details can compromise both clinical relevance and health equity, reinforcing the very disparities such systematic reviews are designed to reduce.

Building on existing work that treats narrative synthesis as a form of multi-document summarisation \citep{deyoung-etal-2024-multi, Wallace2021NarrativeSummaries}, we investigate how well LLMs preserve demographic integrity, specifically age-related descriptors, when approximating a systematic review abstract from the abstracts of included primary studies. 
To support this investigation, we construct DemogSummary, an age-stratified dataset of systematic reviews and primary studies grouped by population, and propose the Demographic Salience Score (DSS), a composite metric that quantifies entity-level retention, omission, and hallucination of age-related information. Using this framework, we assess the summarisation performance of three state-of-the-art LLMs across child, adult, and older adult populations.
\begin{figure*}[ht]  
    \centering
    \includegraphics[width=\textwidth]{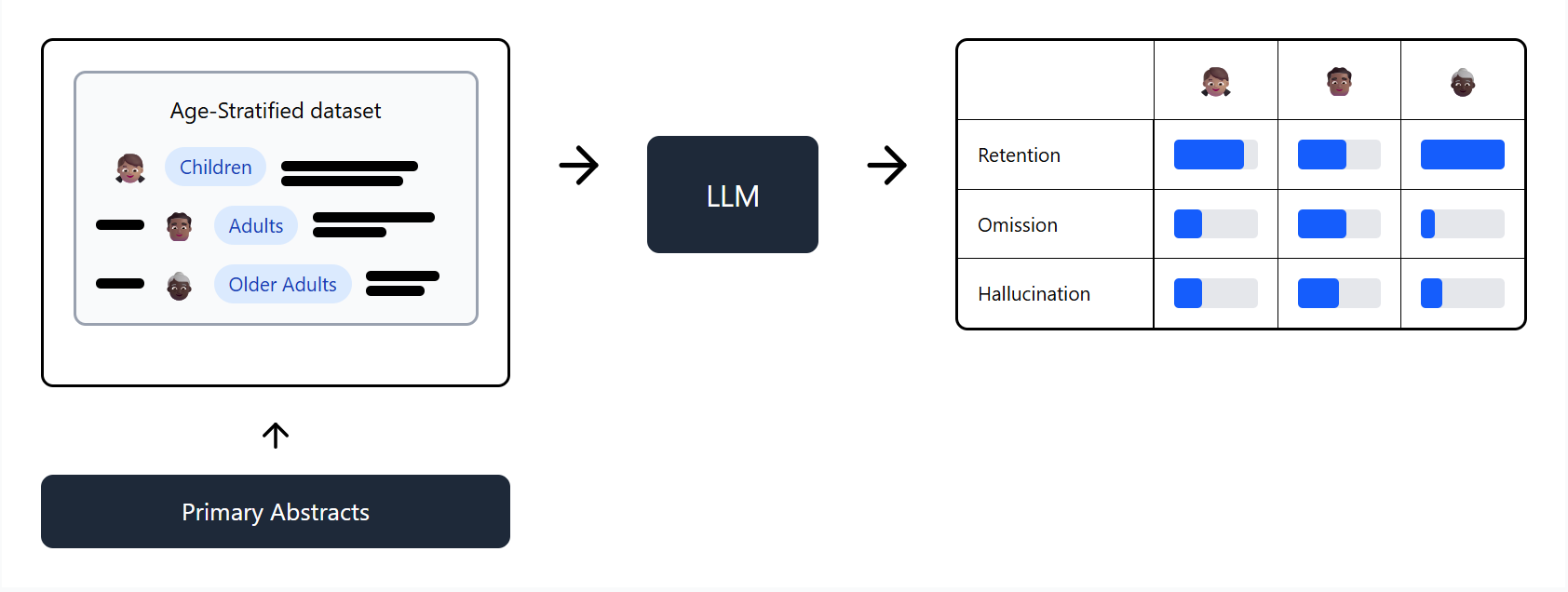}  
    \caption{Age-stratified primary study abstracts are summarised by LLMs, and the outputs are compared to systematic review abstracts. Summaries are evaluated by age group for demographic fidelity using retention, omission, and hallucination metrics.}
    \label{Figure 2}  
\end{figure*}
Our results reveal that fidelity to demographic information is not uniform. Summaries concerning adult populations show the lowest retention of age-related content and the highest incidence of hallucinated descriptors, while those focused on children and older adults show greater accuracy. Across models, GPT-4.1 Nano \citep{openai2023gpt4} and Longformer \citep{beltagy2020longformerlongdocumenttransformer} demonstrate stronger demographic preservation than Qwen-2.5 \citep{yang2025qwen251mtechnicalreport}, though all exhibit limitations in faithfully synthesising age-specific biomedical content. These findings highlight the need to move beyond generic evaluation metrics and toward assessments capturing demographic fidelity in biomedical summarisation. {Our contributions are threefold:}
\begin{itemize}
\item Presenting a novel age-stratified dataset of systematic review primary studies (\textbf{DemogSummary}), grouped by population (children, adults, older adults), designed to support demographic-specific evaluation of summarisation models in biomedical domains.
\item Identifying systematic disparities in how LLMs preserve age-related information during multi-document summarisation, highlighting representational gaps that are obscured by conventional evaluation metrics.
\item Introducing the \textit{Demographic Salience Score}, a targeted metric that quantifies the retention, omission, and hallucination of demographic entities, enabling a fairness-aware assessment of summarisation fidelity.
\end{itemize}
\section{Related Work}
\subsection{Automatic Systematic Reviews}
Data extraction and synthesis represent the most resource-intensive and error-prone phases within the systematic review workflow, often requiring significant manual effort and rigour. Surveys of professionals involved in systematic reviews reinforce this notion; 45\% of respondents in one study of 194 professionals identified data extraction as the most time-consuming stage \citep{Scott2021}. Consequently, there has been an increased focus on automating these steps, particularly using LLMs \citep{ge_leveraging_2024,schmidt_data_2023-1}.
While LLMs have shown promise in biomedical information extraction, their accuracy remains inconsistent across tasks and domains. More broadly, concerns have emerged about their tendency to amplify societal biases, especially in high-stakes fields like medicine. Biases in training data, model design, and linguistic priors can lead to unequal treatment of groups based on attributes such as race \citep{yang_unmasking_2024} and gender \citep{bajaj-etal-2024-evaluating,tang_gendercare_2024}, resulting in harms like stereotyping and misassociation \citep{gallegos_bias_2024}. Yet demographic attributes like \textit{age} remain underexplored in bias evaluations, particularly in biomedical synthesis.
Automatic summarisation selects and integrates key content from one or more documents to produce a concise, informative output \citep{nenkova_automatic_2011}. This task is increasingly important in the biomedical domain, where the exponential growth of publications makes manual synthesis difficult to scale \citep{pawar_survey_2023}. In particular, multi-document summarisation offers a promising way to approximate the narrative synthesis found in systematic reviews. As LLMs are increasingly adopted for this purpose, it becomes essential to examine not only their efficiency, but also how reliably and equitably they preserve critical population-specific information.

\subsection{Bias in Automated Synthesis and Summarisation}
Despite the promise of LLMs for multi-document biomedical summarisation, their outputs often reflect biases that can distort or omit critical demographic information. This is especially concerning in systematic reviews, where under-representation or misrepresentation of population groups may have direct clinical and policy implications.
Prior work has examined bias in summarisation across domains such as news \citep{steen-markert-2024-bias}, opinion generation \citep{huang-etal-2023-examining, huang-etal-2024-bias}, and radiology reports \citep{Seyyed-Kalantari2021, doi:10.1148/radiol.232286}. However, biomedicine has received little attention, particularly in narrative synthesis of systematic reviews, where omissions or hallucinations of population-specific information can compromise the validity of clinical evidence. 
Fairness and bias have been long-standing concerns in NLP \citep{bolukbasi_man_2016, bender_data_2018, blodgett-etal-2020-language, tang_gendercare_2024}, though their definitions are often inconsistent across tasks and domains. We draw on the framework posited by \citet{crawford_trouble_2017}, which distinguishes between two forms of algorithmic bias: allocative bias, which affects access to resources, and representational bias, which misrepresents or excludes certain groups \citep{sun_mitigating_2019, suresh_framework_2021}. This study focuses on representational bias, particularly the loss or distortion of age-related information in generated summaries. We adopt the definition of an unbiased summariser proposed by \citet{steen_bias_2024}, which emphasises faithful and complete representation of relevant input content.

Evaluating bias in summarisation presents several challenges. In certain domains, researchers can generate synthetic or controlled input texts to isolate the effects of bias in model outputs. In biomedical summarisation, however, such input manipulation is impractical, as real clinical studies must be used to reflect the actual conditions and constraints of systematic reviews. To address this, we curate \textbf{DemogSummary}, a real-world, age-stratified dataset that allows for population-specific evaluation without altering the original inputs. This design lets us examine representational bias as it occurs naturally, while controlling for demographic focus through corpus structure and targeted evaluation. Another challenge lies in how summaries are evaluated: standard automatic metrics such as BLEU \citep{papineni-etal-2002-bleu} and BERTScore \citep{DBLP:journals/corr/abs-1904-09675} measure surface or semantic similarity but fail to capture representational distortions or omissions \citep{deutsch-etal-2022-examining, gao-wan-2022-dialsummeval}.
To address this limitation, we complement existing metrics with a novel evaluation method: the \textit{Demographic Salience Score}, which quantifies how well age-related demographic entities are preserved in generated summaries. This allows us to assess summarisation fidelity through the lens of demographic representation, which is an essential but often overlooked dimension in biomedical NLP.
\section{The \textsc{DemogSummary} Dataset}
Existing systematic review datasets, such as Synergy \citep{HE6NAQ_2023}, do not support stratification by population age group, which is essential for our analysis. We therefore construct a new dataset that enables explicit categorisation of primary studies and reviews into child, adult, and older adult populations.
\subsection{Dataset Construction}
Systematic reviews were selected based on the presence of demographic terms (e.g., `Aged`, `Older Adult`, `Adult`, `Child`) in titles or Medical Subject Headings (MeSH) annotations. Searches were conducted in three open-access biomedical databases: PubMed \citep{ncbi2025}, Cochrane \citep{cochrane2025}, and Web of Science \citep{clarivate2025}, using combinations of `systematic review` and demographic keywords. This yielded reviews across medical domains, categorised into three population groups: children (aged <18 years), adults (aged 28-59 years), and older adults (60+ years). Inclusion required that reviews (i) focus on a single, well-defined demographic group and (ii) cite at least three primary studies with accessible abstracts. Any reviews that did not comply with (i) and (ii) were excluded. For each included review, cited primary studies were retrieved via PubMed and PubMed Central identifiers (PMIDs/PMCIDs); if unavailable, in-text hyperlinks were used. The final dataset qualifies under the \citet{ukcdpa1988_29a} exemption for non-commercial text and data analysis\footnote{We release the PubMed IDs for systematic reviews in \textsc{DemogSummary} alongside the code at: \href{https://github.com/Favour-Yahdii/lllms\_dont\_see\_age}{https://github.com/Favour-Yahdii/lllms\_dont\_see\_age} to support transparency, reproducibility, and future research.}.
\subsection{Demographic Annotation}
We identified demographic information using a combination of rule-based methods and LLM-based named entity recognition; details are included in Appendix \ref{dem-annote}.
\begin{table}[ht]
\centering
\footnotesize
\setlength{\tabcolsep}{6pt} 
\begin{tabular}{lcc}
\toprule
\textbf{Age Group} & \textbf{Reviews} & \textbf{Avg. Studies per Review} \\
\midrule
Adults & 14 & 23 \\
Children & 15 & 32 \\
Older Adults & 15 & 25 \\
\bottomrule
\end{tabular}
\caption{Overview of the DemogSummary Dataset. }
\label{tab:dataset}
\end{table}
\begin{table}[ht]
\centering
\footnotesize
\setlength{\tabcolsep}{6pt} 
\begin{tabular}{lcc}
\toprule
\textbf{Medical Domain} & \textbf{Number of Reviews} \\
\midrule
Public Health & 14\\
Frailty & 2\\
Mental Health & 8 \\
Nutrition and Digestive Health & 9 \\
Cognitive Health & 1 \\
Cardiovascular Health & 5 \\
Pain &  2  \\
Dental and Gut Health & 2 \\
Respiratory Health & 1 \\
\midrule
Total & 44 \\
\bottomrule
\end{tabular}
\caption{Breakdown of the \textsc{DemogSummary} Dataset by Medical Domain. }
\label{tab:dataset_domain}
\end{table}
The resulting dataset in Table~\ref{tab:dataset} includes 14 adult, 15 child, and 15 older adult systematic reviews, each comprising primary studies spanning multiple medical domains as described in Table ~\ref{tab:dataset_domain}. In total, these reviews encompass approximately 1,200 primary studies. Although the dataset consists of 44 systematic reviews, this scale is appropriate given the nature of systematic reviews, which are intended to synthesise large bodies of evidence and typically include many primary studies each.
\section{Experimental Design}
\subsection{Model Selection}
We selected three large language models that reflect a range of architectures, context-handling capabilities, and access modalities: 
\paragraph{QWEN} (Qwen/Qwen2.5-14B-Instruct-1M) \citep{yang2025qwen251mtechnicalreport}: a state-of-the-art open-source autoregressive transformer model with extended context length support. QWEN was selected due to its ability to process long sequences (up to 1M tokens), which is critical for working with multiple primary study abstracts without truncation.
\paragraph{GPT} (OpenAI/gpt-4.1-nano) \citep{openai2023gpt4}: a proprietary model accessed via the OpenAI API. We included this model as a strong commercial baseline, selected for its favourable trade-off between speed and performance in general-purpose language understanding tasks. 
\paragraph{Longformer} (allenai/led-large-16384-arxiv) \citep{beltagy2020longformerlongdocumenttransformer}: a transformer-based encoder-decoder model pretrained on scientific papers from ArXiv and designed specifically for scientific long-document processing, making it particularly suitable for tasks involving structured academic language and domain-specific content. 

\subsection{Task Setup}
Each model is tasked with generating an abstractive narrative synthesis based on the full set of primary study abstracts from a systematic review. The target output is a concise summary approximating the published systematic review abstract. We evaluated two prompt conditions: a \emph{regular prompt}, which did not refer to patient demographics, and an \emph{age-aware prompt}, which explicitly informed the model of the population group involved. Both prompts are outlined in (Table \ref{tab:reg_prompt}). This setup enabled assessment of whether demographic cues influence summarisation behaviour.

\subsection{Implementation Details} \label{sec:appendix-taskdetails}
All experiments were conducted over approximately 90 GPU hours on a single NVIDIA L4. Inference with GPT-4.1-nano via the OpenAI API cost approximately \$5 for all reviews, while the open-source models (QWEN and Longformer) were accessed via Hugging Face. More details, including runtime environment, inference cost, and hyperparameters, are provided in Appendix ~\ref{hyperparameters}. 

\begin{table}[!htb]
\centering
\resizebox{0.8\columnwidth}{!}{
\footnotesize
\begin{tabular}{|p{7cm}|}
\hline
\rule{0pt}{1.1em}
\noindent You are an experienced and objective biomedical systematic reviewer.
Your task is to draft a concise, structured abstract that approximates a systematic review abstract, using the set of provided biomedical research abstracts.\textbf{The provided research abstracts involve studies conducted specifically in \{POPULATION GROUP\} populations. Keep this in mind as you complete the task.}
Ensure to produce your summary abstract based only on the provided abstracts. Do not include any external information or personal opinions. Your summary should be a synthesis of the provided abstracts, not a critique or evaluation of them. \\
\hline
\end{tabular}}
\caption{Prompt used for summary generation. The text in \textbf{bold} was included only in the age-aware prompt.}
\label{tab:reg_prompt}
\end{table}

\section{Evaluation Metrics}
We conducted a supervised evaluation using the published systematic review abstracts as gold standards. In addition to standard summarisation metrics, we introduced the \textit{Demographic Salience Score} to assess retention of demographic content.
\subsection{Standardised Metrics}
\label{sect:standard}
We employed a suite of complementary evaluation metrics to capture different aspects of summarisation quality. BLEU \citep{papineni-etal-2002-bleu} measures surface-level n-gram overlap with the gold-standard abstract, while BERTScore \citep{DBLP:journals/corr/abs-1904-09675} evaluates semantic similarity using contextual embeddings from a pre-trained transformer.
To assess fluency, informativeness, and alignment with expert-authored content, we used BARTScore \citep{NEURIPS2021_e4d2b6e6} in a supervised setup, where the model computes the likelihood of the generated summary given the gold standard. 
Lastly, FactCC \citep{kryscinski-etal-2020-evaluating} was used to assess factual consistency. It classifies each sentence in the generated summary by whether it is supported by the gold summary or not. 

\subsection{Demographic Salience Score (DSS)}
To evaluate how effectively demographic information is preserved and conveyed in multi-document summarisation, we introduce the Demographic Salience Score (DSS). While this study focuses on age-related demographic content, the metric is generalisable to other demographic dimensions (e.g., gender, ethnicity). DSS captures two key aspects of demographic fidelity: (1) inclusion of salient demographic entities, and (2) penalisation of unsupported (hallucinated) content.  

\paragraph{Entity Extraction.}
We construct a gold-standard set of demographic entities by parsing each systematic review. Entities related to age are extracted using a combination of rule-based patterns and LLM-assisted methods\footnote{Details of these techniques are provided in Appendix \ref{dem-annote}.}. These extracted entities form the reference set $Ent_{gold}$, which serve as the evaluation target.

To evaluate the reliability of the demographic annotation process, we manually reviewed a stratified random sample of 60 annotated abstracts, 20 from each age group (children, adults, older adults), representing approximately 5\% of the total dataset.

In this subset, 59 out of 60 annotations (98\%) were accurate.  Based on this high accuracy rate and the diversity of the sample, we considered the annotation pipeline to be sufficiently reliable for downstream analysis.

Each primary study was subsequently categorised into one of the following age groups; \textbf{Child}: <18 years, \textbf{Adult}: 18–59 years, \textbf{Older Adult}: 60+ years.
\begin{figure}[ht]  
    \centering
    \includegraphics[width= 0.8\columnwidth]{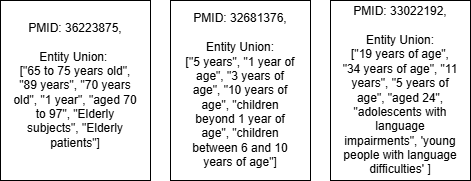}  
    \caption{Subset of Gold standard Demographic Entities from Reviews}
    \label{Figure 1}  
\end{figure}


\paragraph{Scoring Components.}
The DSS has two main components, the Entity Retention Score (ERS) and the Hallucination Penalty (HP). Components such as the omission rate are derivatives of these main components. We establish $Ent_{summary}$ as the set of demographic entities in the generated summary.
\paragraph{Entity Retention Score (ERS):} The proportion of gold entities preserved in the generated summary. This score reflects how comprehensively the summary captures the reference demographic content.
\begin{equation}
    \label{eq:ERS}
    ERS = \frac{|Ent_\text{summary} \cap Ent_\text{gold}|}{|Ent_\text{gold}|}
\end{equation}

\textit{Omission Rate.} As the complement of ERS, we define omission as the proportion of gold entities missing from the summary:

\begin{equation}
\label{omission_eqn}
Omission = 1 - ERS
\end{equation}

Although not part of the DSS formulation, we report omission rates in our results to provide an additional perspective on demographic coverage.

\paragraph{Hallucination Penalty (HP):} The proportion of entities in $Ent_{summary}$ that do not match any entity in $Ent_{gold}$. A generated entity is considered a match if it either exactly matches a gold entity or exceeds a specified cosine similarity threshold\footnote{see Section \ref{implementation_deets} for details}.

\begin{equation}
  \label{eq:HP}
  HP = \frac{|Ent_\text{summary} \setminus \text{Match}(Ent_\text{gold})|}{|Ent_\text{summary}|}
\end{equation}

\textit{Over-length Penalty (OP):}  In multi-document summarisation, excessively long outputs can artificially deflate the hallucination penalty by increasing the total number of extracted entities—thereby reducing the proportion of hallucinated content. To account for this, we introduce an over-length penalty (OP) that activates when the number of generated tokens exceeds a predefined threshold. This adjustment ensures that the hallucination penalty remains sensitive to unsupported content, regardless of summary length:

\begin{equation}
OP = \frac{\max(0, T_{\text{gen}} - T_{\text{max}})}{T_{\text{max}}}
\label{eq:overpenalty}
\end{equation}

where $T_{\text{gen}}$ is the number of generated tokens, and $T_{\text{max}}$ is a predefined token limit.
The adjusted hallucination penalty is then defined as:

\begin{equation}
HP_{adj} = HP + OP
\end{equation}

\paragraph{DSS:} The Demographic Salience Score is computed as a weighted combination of retention and penalty terms:

\begin{equation}
    \label{eq:dss-final}
    \text{DSS} = \alpha \times \text{ERS} - \gamma \times \text{HP}_{\text{adj}}
\end{equation}

Where $\alpha$, and $\gamma$ are non-negative weighting coefficients.
To normalise the score to the $[0, 1]$ range, we divide by the maximum achievable score $\alpha \cdot N$ (i.e., a perfect score with full retention and no hallucination penalty), and clip negative values, where $N$ is the number of systematic reviews:
\begin{equation}
    \label{eq:DSS-Norm}
    \text{DSS}_{\text{normalised}} = \max\left(0, \frac{\alpha \times \text{ERS} - \gamma \times \text{HP}_{\text{adj}}}{\alpha \times N} \right)
\end{equation}
This normalisation ensures interpretability while bounding the score, rewarding summaries that retain salient demographic content and penalising unsupported or excessive additions.
\paragraph{Implementation Details.}
\label{implementation_deets}
We identified matched demographic entities using cosine-based semantic similarity, applying a threshold of 0.7 to determine matches\footnote{We also explored prompting an LLM for entity matching but found it inconsistent and less reproducible.}. The same similarity threshold was applied inversely to identify hallucinations,i.e., entities with similarity below the threshold were considered unsupported. Weighting parameters $\alpha=\gamma=2$ were used in Equations~\ref{eq:dss-final} and~\ref{eq:DSS-Norm}, balancing the contribution of entity retention and hallucination penalties. These settings were chosen to prioritise demographic salience while discouraging unsupported content \footnote{We performed sensitivity analysis tests for these parameters; details can be found in  Appendix ~\ref{sec:appendix-sensitivity analysis}.}.

\paragraph{Metric Comparison and Interpretation.}
To contextualise the behaviour of DSS, we examined its relationship with the standard evaluation metrics introduced in Section~\ref{sect:standard}. Pearson correlation analysis showed strong positive correlations between DSS and BLEU (0.970), BERTScore (0.923), and BARTScore (0.892), indicating that summaries with higher demographic fidelity often also exhibit strong lexical and semantic alignment with reference abstracts.
However, despite this alignment, DSS remains conceptually distinct. Unlike standard metrics, which assess surface-level similarity, DSS explicitly measures the retention of demographic entities and penalises unsupported or omitted demographic information. In contrast, metrics such as BERTScore do not distinguish between which information is preserved or hallucinated, nor do they prioritise content relevance tied to specific population groups.

DSS also showed a negative correlation with FactCC (-0.99), a factual consistency metric, suggesting that DSS captures a complementary dimension of quality, namely, demographic salience and inclusion, that is not the primary focus of factuality-oriented metrics. Taken together, these comparisons indicate that DSS aligns with general measures of summary quality but contributes a focused, domain-relevant perspective that standard metrics do not directly capture.

\subsection{Model-Level Performance Analysis}
To assess differences in model performance across the three age groups, we used the Kruskal–Wallis test \citep{ref1}, a non-parametric method appropriate for comparing three or more independent groups without assuming normality. When significant effects were found, Dunn’s post-hoc tests \citep{dunn} with Bonferroni correction were applied to adjust for multiple comparisons. Effect sizes were calculated using epsilon-squared ($ \varepsilon^2 $).

\section{Results and Discussions}
\begin{table*}[ht]
\centering
\resizebox{\textwidth}{!}{
\footnotesize
\begin{tabular}{llccccccccc}
\toprule
\textbf{Group} & \textbf{Model} & \textbf{BERT} $\uparrow$ & \textbf{BART} $\uparrow$ & \textbf{BLEU} $\uparrow$ & \textbf{FactCC} $\uparrow$ & \textbf{ERS} $\uparrow$ & \textbf{HP} $\downarrow$ & \textbf{Omission} $\downarrow$ & \textbf{OP} $\downarrow$ & \textbf{DSS} $\uparrow$ \\
\midrule
\multicolumn{11}{c}{\textbf{Regular Prompt}} \\
\midrule
Child & GPT  &  \textbf{0.84} & \textbf{-2.26} & 2.40 & \textbf{0.76}& 0.84 & \textbf{0.12} & 0.16& \textbf{0.00} & \textbf{0.72}\\
                & QWEN & 0.82 &  \textbf{-2.26} & 1.68 & 0.52&  \textbf{0.97} & 0.58 & \textbf{0.02} & 0.95& 0 (-0.55)\\
                & Longformer  & 0.81 & -2.30 &  \textbf{2.46} & 0.44& 0.91 & 0.33 & 0.09 & \textbf{0} & 0.63\\ \midrule
Adult & GPT   & \textbf{0.83} & -2.31& 2.05& \textbf{0.82}& \textbf{0.81} & \textbf{0.12} &  \textbf{0.19} & \textbf{0} & \textbf{0.69} \\
             & QWEN  & 0.81 & \textbf{-2.30} & 1.76 & 0.51& 0.78 & 0.74 & 0.22&1.02 & 0 (-0.98)\\
            & Longformer & 0.80 & -2.37 & \textbf{2.16} & 0.60& 0.45 & 0.18 & 0.50 & \textbf{0.00} & 0.27\\ \midrule
Older Adult & GPT  & \textbf{0.83} &-2.00  & \textbf{3.75} & \textbf{0.71}& 0.92 & 0.14 & 0.08 & \textbf{0} & 0.78\\
                        & QWEN & 0.82 & \textbf{-1.99}  & 2.25 & 0.43& \textbf{0.98} & 0.11 & \textbf{0.02} & \textbf{0} & \textbf{0.79}\\
            & Longformer  & 0.81 & -2.03 & 2.81 & 0.41& 0.95 & \textbf{0.07} & 0.05 & \textbf{0} & 0.78\\
\midrule
\multicolumn{11}{c}{\textbf{Age-Aware Prompt}} \\
\midrule
Child & GPT &\textbf{0.84} & \textbf{-2.25} & \textbf{3.04} & \textbf{0.83}& 0.87 &  \textbf{0.17}&  0.13 & \textbf{0} & \textbf{0.69}\\
            & QWEN &0.81 & -2.31 & 0.72& 0.63&  0.89 &  2.06&  0.11  & 1.32 &0( -2.49)\\
    & Longformer & 0.82&-2.31 &2.04 & 0.59&  \textbf{0.91} &  0.33 &  \textbf{0.09}&  \textbf{0} & 0.58\\ \midrule
Adult & GPT         &\textbf{0.83} &-2.31 &\textbf{2.60} &  \textbf{0.82}&   0.62 &  \textbf{0.12} &  0.38  & \textbf{0} & \textbf{0.5}\\
             & QWEN   &\textbf{0.83} & \textbf{-2.27} & 1.62 &  0.59&     \textbf{0.93} &   1.14 &  \textbf{0.07} & 1.57 &0(-1.78) \\
        & Longformer & 0.81&-2.35 & 2.18&  0.63&  0.45&  0.18&  0.55 & \textbf{0} & 0.27 \\ \midrule
Older Adult & GPT   &\textbf{0.84} & -2.00 & \textbf{3.70} &  \textbf{0.67}& 0.85 &  \textbf{0.04} &  0.15  & \textbf{0} & 0.81 \\
            & QWEN &0.83 &\textbf{-1.98}&2.27 &  0.42&  \textbf{0.98} &  0.11&  \textbf{0.02} & 1.33 &0(-0.46) \\
    & Longformer &0.81 & -2.04&1.93& 0.50&  0.95&  0.07&  0.05&  \textbf{0} & \textbf{0.88} \\
\bottomrule
\end{tabular}}
\caption{Comprehensive automatic evaluation results across demographic groups and summarisation models.
The top section reports scores under the regular prompt, and the bottom under the age-aware prompt. ERS, HP, Omission and OP are reported as averages across reviews. DSS is reported as the normalised value, with negative unnormalised scores in parentheses.}
\label{tab:full_metric_regular}
\end{table*}
\paragraph{Surface Metrics Are Demographically Insensitive; FactCC Shows Partial Sensitivity.}
Based on the results in Table ~\ref{tab:full_metric_regular}, BLEU, BERTScore, and BARTScore exhibit minimal variation across models and age groups, indicating a lack of sensitivity to population-specific fidelity and limits their utility in demographic evaluation.
While limited, FactCC reveals greater differentiation. GPT consistently scores higher, with age-aware prompts improving performance in the child group. However, scores remain low for older adults across QWEN and Longformer, despite similar gains, indicating ongoing challenges in summarising this group. The divergence between high FactCC scores and poor demographic fidelity underscores the need for metrics that capture representational accuracy.

\paragraph{DSS Highlights Model-Specific Tradeoffs.}
DSS, designed to assess demographic fidelity, reveals clearer distinctions between models. Under regular prompting, GPT achieves high DSS across all groups by balancing entity retention with low hallucination and length penalties. In contrast, QWEN, despite strong ERS, suffers from high hallucination and overly verbose summaries, resulting in sharply reduced DSS, especially in adults and children. Longformer exhibits fewer hallucinations overall but is notably uneven across age groups: it performs relatively well for older adults and children, where omission rates are lower, but struggles with the adult group, where retention is markedly weaker. Overall, in the older adult group, DSS is consistently higher.

\paragraph{Simple Fixes Are Non-Trivial: Age-Aware Prompting Yields Mixed Effects on Fidelity and Stability.} The impact of age-aware prompting varies considerably across models and age groups. For GPT, it leads to modest gains in factual consistency (FactCC) but slightly reduces DSS due to lower entity coverage. QWEN shows high sensitivity to age aware prompting, achieving near-perfect ERS in some settings, but at the cost of increased hallucination and verbosity, leading to substantial drops in DSS. 
Longformer’s behaviour is more consistent: It shows limited improvement overall but performs relatively well in the older adult group, where DSS increases without major penalty tradeoffs. These results suggest that while demographic specific age-aware prompting can shift model behaviour, it does not consistently enhance demographic fidelity and often introduces new sources of instability, thereby highlighting the non-triviality of improving model's retention of salient demographic entities.

\begin{figure*}[ht]
\centering
\footnotesize

\begin{subfigure}[b] {0.45\textwidth}
    \centering
    \includegraphics[width=\linewidth]{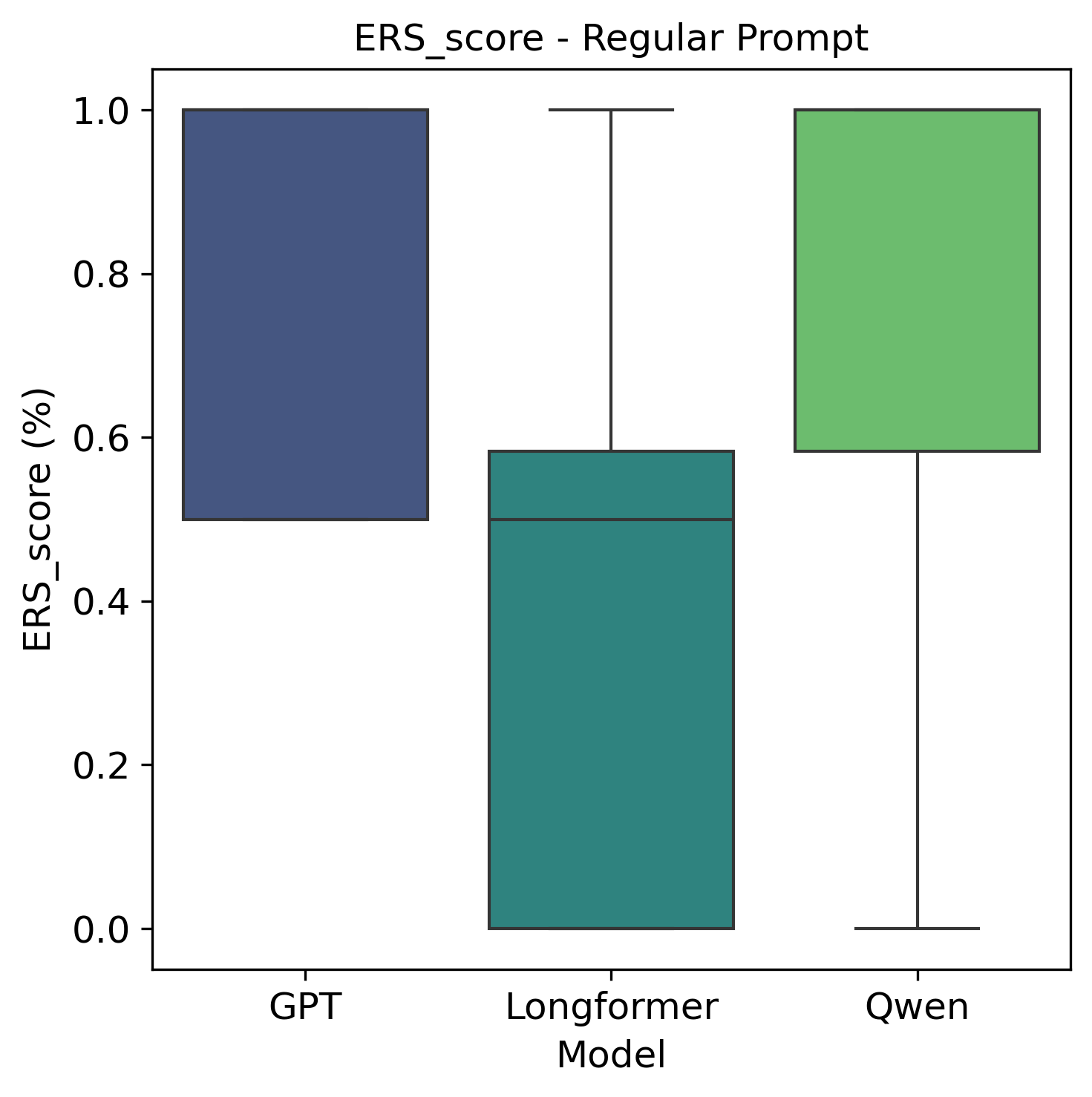}
    \caption{Entity Retention Across Models (Adult Group) - Regular Prompt}
    \label{fig:ers_reg}
\end{subfigure}
\hspace{0.05\textwidth}
\begin{subfigure}[b]{0.45\textwidth}
    \centering
    \includegraphics[width=\linewidth]{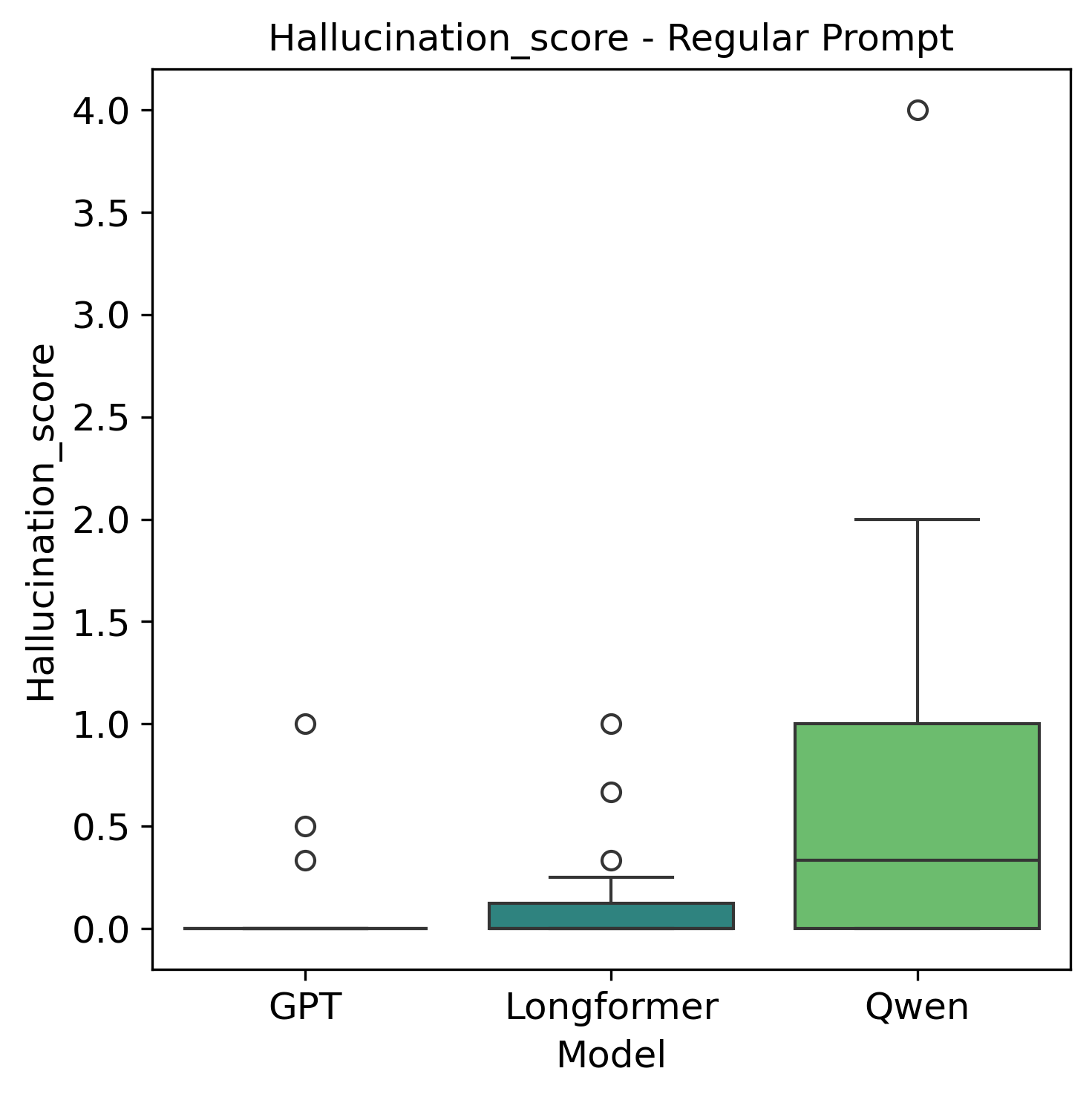}
    \caption{Hallucinations Across Models (Adult Group) - Regular Prompt}
    \label{fig:hal_reg}
\end{subfigure}

\caption{Comparison of Entity Retention and hallucinations across Models for the Adult Group - Regular Prompt.}
\label{fig:adult_side_by_side}
\end{figure*}

\subsection{Retention and Hallucination Patterns Across LLMs.}
While Table~\ref{tab:full_metric_regular} summarises model-level averages, it obscures instance-level variability that can affect the reliability of summarisation in practice. Figure ~\ref{fig:adult_side_by_side} shows the distribution of Entity Retention Score (Figure ~\ref{fig:ers_reg}) and Hallucination Score  (Figure ~\ref{fig:hal_reg}) for the adult group under the regular prompt.
The ERS distribution highlights GPT's consistently strong demographic coverage, with minimal variation across reviews. In contrast, Longformer displays a wide spread and lower median, suggesting inconsistent retention of key population information. QWEN achieves comparable retention but at the cost of highly variable hallucination scores, including extreme outliers, indicating instability in content faithfulness. These patterns suggest that even when models appear comparable on average, their behaviour can diverge substantially across instances, a risk especially critical in biomedical settings. 
Additional distributions for other age groups and prompt conditions are provided in Appendix~\ref{sec: appendix ers}, along with a token-level qualitative analysis in Section ~\ref{sec:qual eval}.

\subsection{Statistical Analysis of Inter-Model Differences}
We assessed whether model-level differences across age groups and evaluation metrics were statistically significant using the Kruskal–Wallis and Dunn’s tests. Results below summarise key findings for each prompt condition.
\paragraph{Regular Prompt.}
Under the regular prompt, significant differences appeared mainly in the adult group. ERS and omission scores differed across models (\( H = 10.30, p = 0.0058, \varepsilon^2 = 0.20 \)), with Longformer underperforming relative to GPT (\( p = 0.013 \)) and QWEN (\( p = 0.021 \)); differences between GPT and QWEN were not significant (\( p = 1.00 \)). Hallucination scores also varied (\( H = 7.23, p = 0.0269, \varepsilon^2 = 0.12 \)), with QWEN exceeding GPT (\( p = 0.042 \)).
In the child group, only hallucination differences were significant (\( H = 6.14, p = 0.046, \varepsilon^2 = 0.10 \)), with QWEN hallucinating more than GPT (\( p = 0.040 \)). No significant differences were found in the older adult group (\( p > 0.90 \)).
\paragraph{Age-Aware Prompt.}
With the age-aware prompt, inter-model differences became more pronounced for adults. ERS and omission scores again differed significantly (\( H = 11.40, p = 0.0034, \varepsilon^2 = 0.22 \)), with Longformer performing worse than QWEN (\( p = 0.0027 \)); GPT and QWEN did not differ significantly (\( p = 0.088 \)). Hallucination scores also varied (\( H = 13.10, p = 0.0014, \varepsilon^2 = 0.26 \)), with QWEN exceeding both GPT (\( p = 0.0023 \)) and Longformer (\( p = 0.013 \)).
Among children, hallucination scores showed strong differences (\( H = 19.27, p = 6.54 \times 10^{-5}, \varepsilon^2 = 0.41 \)), with QWEN again exceeding GPT (\( p = 0.00012 \)) and Longformer (\( p = 0.0020 \)); ERS and omission were not significant. Older adults showed no significant differences on any metric (\( p > 0.10 \)).

Overall, model variability was most pronounced in adults, particularly for ERS and hallucinations. QWEN consistently hallucinated more than GPT and Longformer, while Longformer’s lower ERS and higher omission were concentrated in the adult group.

\subsection{Qualitative Evaluation and Token Analysis}
\label{sec:qual eval}
Entity-level behaviour was examined across three dimensions: entities retained, hallucinations, and omissions. These were evaluated across demographic groups and under both regular and age-aware prompts. 
\paragraph{Entity Retention.}
Retained entities spanned a mix of canonical age-group labels (e.g., \emph{adults}, \emph{children}, \emph{older adults}) and more granular descriptors such as \emph{young adults}, \emph{adults aged 18--35}, \emph{early childhood}, and \emph{community-dwelling older adults}. The age-aware prompt tended to elicit more specific and contextually rich terms, including detailed demographic subgroups (e.g., \emph{healthy elderly men and women}, \emph{Midwestern young adults}, \emph{adult cardiac patients}). These findings suggest that models were generally sensitive to age-related cues and capable of surfacing relevant synonyms and descriptive variants, especially when the prompt was explicitly age-conditioned.
\paragraph{Hallucinations and Omissions.}
Hallucinated entities, while less frequent overall, often involved tangential or demographically unrelated populations not mentioned in the gold document. Examples include the inclusion of \emph{prisoners} in the population group of a review on smoking cessation interventions for young adults. Hallucination frequency was higher for QWEN and more prominent under the age-aware prompt, reflecting a tendency to overgenerate plausible but unsupported population descriptors. These patterns mirror the quantitative hallucination scores and indicate model-specific susceptibility to fabrication and generation of unsupported content, particularly when prompted to focus on demographic detail.
Omissions included both general and highly specific age-related terms that were present in the gold documents but not captured in the generated outputs. These, similar to retained entities, ranged from descriptive phrases (e.g., \emph{mean age 73.66 $\pm$ 14.67 years}, \emph{residents in aged care settings}) to developmental or life stage terms such as \emph{infancy}, \emph{childhood}, and \emph{young adulthood}. Across prompts, omissions were most pronounced for Longformer, consistent with its lower quantitative entity retention scores. The omission of precise subgroups may reflect limitations in entity grounding, particularly for complex or compound descriptors. There seemed to be no obvious pattern in the models decision on which entities to retain and which to exclude.

\subsection{Effect of Prompt Design}
The age-aware prompt systematically increased the specificity and range of demographic terms produced, particularly for adults and older populations. However, this increase in granularity was accompanied by greater lexical variability and, for QWEN in particular, more hallucinations and unsupported entities. GPT retained a better balance between fidelity and specificity, while Longformer exhibited lower entity coverage across conditions.

Overall, the age-aware prompt improved coverage of fine-grained descriptors but also amplified model-specific failure modes, such as hallucinations or omissions, highlighting the non-triviality of improving demographic entity retention in LLMs. These qualitative observations support the quantitative findings, particularly regarding QWEN's hallucination rate and Longformer's lower entity retention performance.

\section{Conclusion}
Our study shows that current LLMs exhibit systematic disparities in age-related information retention in biomedical summaries. Using \textbf{DemogSummary} and the \textbf{Demographic Salience Score (DSS)}, we quantify these biases, finding adult summaries particularly error-prone and underrepresented populations more likely to be hallucinated. These results highlight the need for demographic-aware evaluation and fair summarisation pipelines, paving the way for more equitable and transparent biomedical NLP systems.
\section*{Limitations}
This study offers a focused evaluation of age-related fairness in LLM-based summarisation but has several limitations. First, the dataset covers a limited set of medical domains, which may constrain generalisability. Second, the causes of observed disparities remain under explored. Lastly, the study does not investigate bias mitigation methods, limiting its prescriptive value for fairness-oriented applications.



\section*{Acknowledgements}
This work was supported by the University of Sheffield's Healthy Lifespan Institute Research Scholarship. We also acknowledge the University of Sheffield IT Services for providing access to High Performance Computing resources. We are grateful to the reviewers and area chairs for their thoughtful feedback and constructive discussions, and especially to Jasivan Sivakumar and Joe Stacey for their invaluable insights and guidance.
\bibliography{latex/custom}

\appendix
\section{Hyperparameters}
\label{hyperparameters}
The hyperparameters in Table ~\ref{tab:hyperparameters} were selected to encourage highly deterministic, concise outputs (temperature = 0) and to reduce redundancy. The maximum output token limit (750) was chosen both to provide sufficient token space for generating detailed summaries that approximate the length and content of gold-standard review abstracts, as well as a guide for penalising overly lengthy syntheses that provide weak and shallow aggregations of biomedical systematic review abstracts.

\begin{table}[h!]
\centering
\footnotesize
\begin{tabular}{lccc}
\hline
\rule{0pt}{1.1em}
\textbf{Model} & \textbf{Temperature} & \textbf{Max Tokens} & \textbf{FP} \\
\hline
GPT-4.1 Mini & 0 & 750 & NA \\

GPT-4.1 Nano & 0 & 750 & NA \\

Qwen & 0 & 750 & 1.1 \\

Longformer & NA & 750 & 1.1 \\
\hline
\end{tabular}
\caption{Hyperparameters used for different models in the experiment. FP = Frequency Penalty}
\label{tab:hyperparameters}
\end{table}

\section{Demographic Annotation Procedure}
\label{dem-annote}
\subsection{Rule-Based Age Entity Extraction}

\begin{lstlisting}[language=Python, caption={Pseudocode for rule-based age extraction}, label={lst:age_extraction}]
def extract_demographics(text):
    demographics = {}

    # Define regular expression patterns to identify age-related expressions
    age_patterns = [
        # Matches "45 years old", "30-50 years"
        
        r'(?<!\d)(\d{1,3})\s*(?:(?:-|to)\s*(\d{1,3}))?\s*(?:years?|yrs?)\s*(?:old|of age)?',
        
        # Matches "aged 40 to 65", "aged 70"
        r'aged?\s*(\d{1,3})(?:\s*(?:to|-|-)s*(\d{1,3}))?',
        # Matches "6 month-old", "12 mo old"
        r'(\d{1,2})\s*(?:month[- ]old|mo old)',
        
        # Matches "age of 20 to 40", "aged 60-85"
        r'(?:age|aged)\s*(?:of\s*)?(\d{1,3})\s*(?:to|-|-)|s*(\d{1,3})'
    ]

    def find_matches(patterns):
        return [match.group().strip() for p in patterns for match in re.finditer(p, text, re.IGNORECASE)]

    demographics["age"] = find_matches(age_patterns)
    return demographics
\end{lstlisting}

\subsection{LLM-Based Named Entity Recognition Prompt}

\begin{lstlisting}[language=Python, caption={LLM prompt for structured demographic extraction}, label={lst:llm_prompt}]
You are an intelligent assistant tasked with extracting structured information from academic PDF documents.

Step 1: Extract the following fields:
- "title"
- "firstauthor"
- "year"
- "abstract" (until the "Keywords" section)
- "systematicreviewpmid"
- "stype"
- "populationdemographics"

Step 2: Return a JSON object in the following format:

{
  "title": "...",
  "firstauthor": "...",
  "year": "...",
  "abstract": "...",
  "systematicreviewpmid": "...",
  "stype": "...",
  "populationdemographics": [...]
}

Notes:
- Read the full abstract across chunks if needed.
- Do not stop early or include extra commentary.
\end{lstlisting}

\section{Sensitivity Analysis of DSS Parameters} \label{sec:appendix-sensitivity analysis}
To evaluate the robustness of the Demographic Salience Score (DSS), we conducted a sensitivity analysis on key hyperparameters: (i) the semantic similarity and hallucination thresholds, and (ii) the weighting parameters $\alpha$ (ERS weight) and $\gamma$ (hallucination penalty).

Threshold analysis showed a consistent trend: DSS scores were highest at lower, more permissive thresholds and declined gradually as thresholds increased, reflecting stricter evaluation. We selected 0.7 for both thresholds, yielding a normalized DSS of 0.64—a balance between semantic precision and penalization of unsupported content. DSS varied minimally in the surrounding parameter space, indicating local stability.

For $\alpha$ and $\gamma$, we tested values between 1 and 2. We avoided assigning 0 to any of the parameters to avoid cancelling the contribution of the corresponding component, which is undesirable. In our tests, increasing $\alpha$ slightly improved DSS, whereas higher $\gamma$ values reduced it due to stronger hallucination penalties. For experimentation, researchers may adjust these weights depending on which aspect they wish to emphasise, assigning higher weight to $\gamma$ to prioritise hallucination control, or to $\alpha$ to strengthen DSS. The final configuration ($\alpha = 2.0$, $\gamma = 2.0$) achieved a DSS of 0.66, one of the highest observed, and exhibited robustness to parameter variation.

\begin{figure*}[ht]  
\footnotesize
    \centering
    \includegraphics[width=0.8\textwidth]{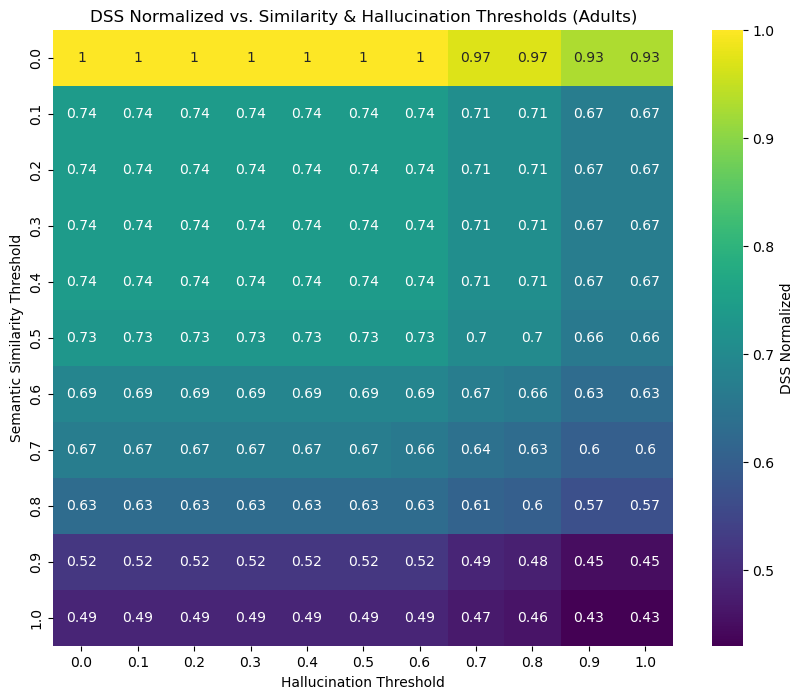}  
    \caption{Semantic Similarity Threshold}
    \label{Figure semantic sim}  
\end{figure*}

\begin{figure*}[ht]  
\footnotesize
    \centering
    \includegraphics[width=0.8\textwidth]{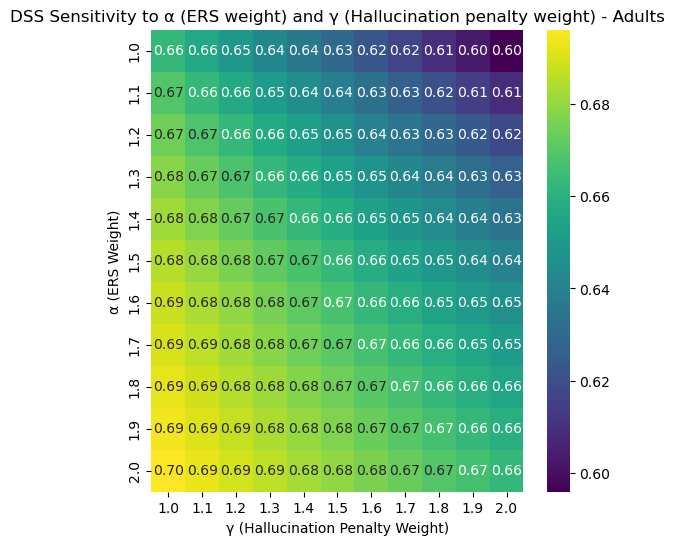}  
    \caption{Alpha-Gamma Grid}
    \label{Figure alpha sim}  
\end{figure*}
Figure ~\ref{Figure alpha sim} displays normalised DSS scores over a grid of semantic similarity and hallucination thresholds, while figure ~\ref{Figure alpha sim} shows DSS values as a function of $\alpha$ and $\gamma$, varied from 1.0 to 2.0. 

\section{Source-Level Demographic Entity Extraction}
\label{sec:appendix}

\begin{lstlisting}[language=Python, caption={Pseudocode for Demographic Extraction Function}]
Function extract_demographics(text):

    Initialise an empty dictionary: demographics

    Define regex patterns for:
        - Age (e.g., "45 years old", "aged 40 to 65", "6 month-old")

    Define helper function find_matches(patterns):
        For each pattern in patterns:
            Use case-insensitive regex search on text
            Collect all matching substrings
        Return list of matches

    demographics["age"] = find_matches(age_patterns)

    Return demographics
\end{lstlisting}

\begin{lstlisting}[language=Python, caption={Pseudocode for Age Entity Extraction}]
Function extract_entities(text):

    Define system prompt:
        "You are a helpful assistant. Given the abstract, extract all age related demographic entities.
        You should extract entities related to age.
        Your job is to extract these entities only, do not add to or subtract from the provided text."

    Send prompt and input text to language model (e.g., GPT):
        - Model: "gpt-4.1-nano"
        - Instructions: system prompt
        - Input: "Here is the abstract set: \n{text}"

    Receive response from model
    Return the output text as extracted entities
\end{lstlisting}

\section{Demographic Salience and Entity Retention}
\begin{lstlisting}[language=Python, caption={Pseudocode for Entity Retention Evaluation and DSS Computation}]
Function compute_semantic_similarity_and_ers(
    records, threshold=0.7, hallucination_threshold=0.7,
    alpha=2, gamma=2, overlength_penalty=None):

    Initialise:
        exact_matches := empty dict
        similar_matches := empty dict
        hallucinations := empty dict
        omissions := empty dict
        omission_scores := empty dict
        ers_scores := empty dict
        hallucination_scores := empty dict

    For each record in records:
        reference_entities := list of gold entities
        generated_entities := list of predicted entities
        ID := record identifier

        Compute:
            exact := intersection of reference and generated entities
            embeddings_ref := embeddings for reference_entities
            embeddings_gen := embeddings for generated_entities

        For each reference entity not in exact:
            If cosine similarity with any generated entity >= threshold:
                Append to similar_matches
                Count as similar match

        For each generated entity not in reference_entities:
            If cosine similarity with all reference entities < hallucination_threshold:
                Add to hallucinations

        For each reference entity not matched:
            If cosine similarity with all generated entities < threshold:
                Add to omissions

        Compute:
            omission_score := omitted / reference_entities
            ers := 1 - omission_score
            hallucination_score := hallucinations / reference_entities

        Store scores for ID

    Compute group-level metrics:
        ers_sum := sum of all ERS scores
        hall_sum := sum of all hallucination scores

        If overlength_penalty is provided:
            Add penalties to hall_sum

        DSS := alpha x ers_sum - gamma x hall_sum
        DSS_normalised := max(0, DSS / (alpha x number of records))

    Return:
        exact_matches, similar_matches, hallucinations, hallucination_scores,
        omissions, omission_scores, ers_scores, DSS, DSS_normalised,
        [overlength_penalty if provided]
\end{lstlisting}

\clearpage
\twocolumn[
\section{Entity Retention, Hallucinations and Omissions}
\label{sec: appendix ers}
\begin{center}
Here we present the full set of results across both prompt regimes and demographic groups tested.
\end{center}
\vspace{13em} 
]
\begin{figure}[!htbp]  
    \centering
    \includegraphics[width=0.9\columnwidth]{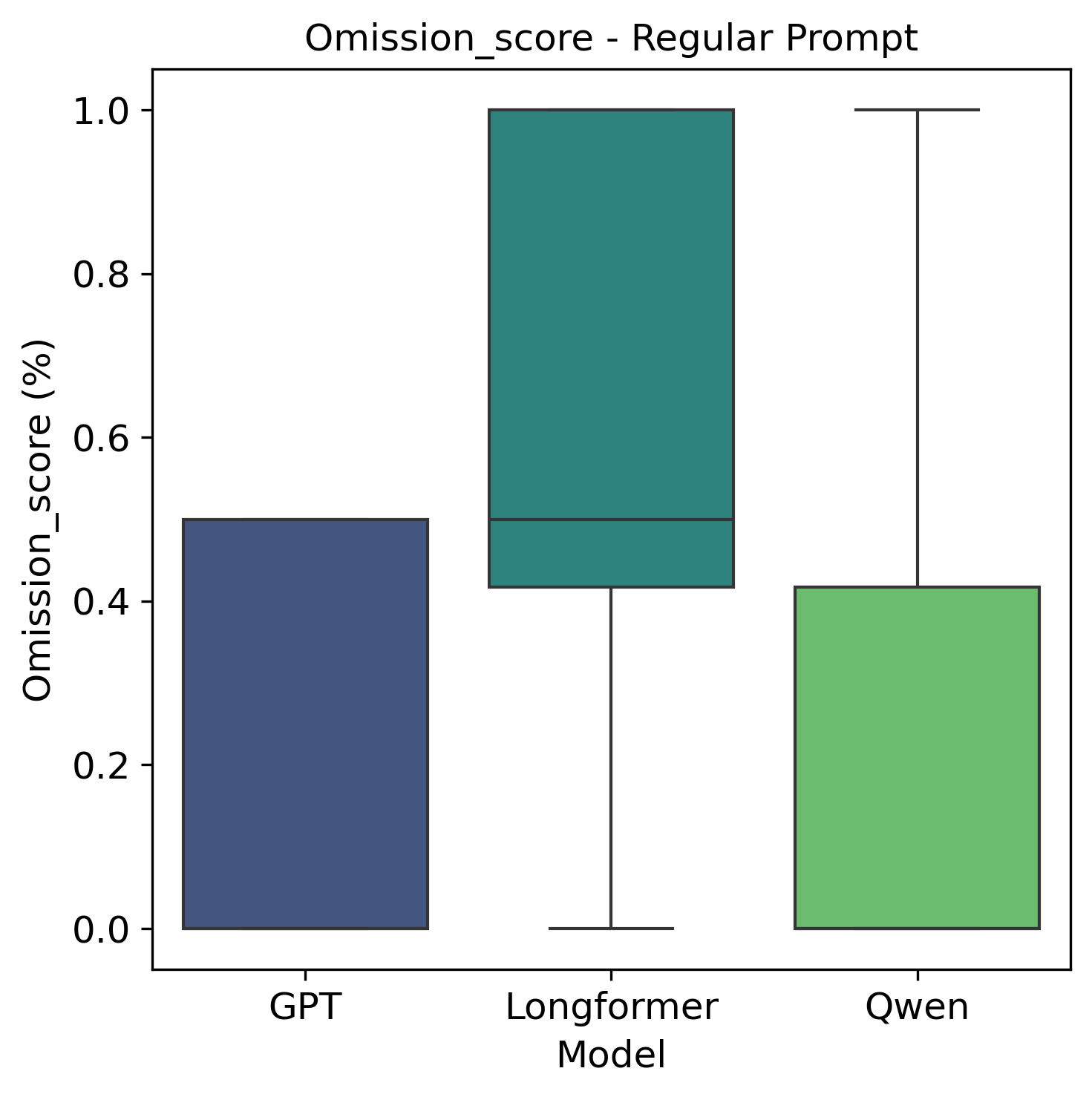}  
    \caption{Entity Omissions Across Models (Adult Group) - Regular Prompt}
    \label{omission reg}  
\end{figure}

\begin{figure}[!htbp]  
    \centering
    \includegraphics[width=0.9\columnwidth]{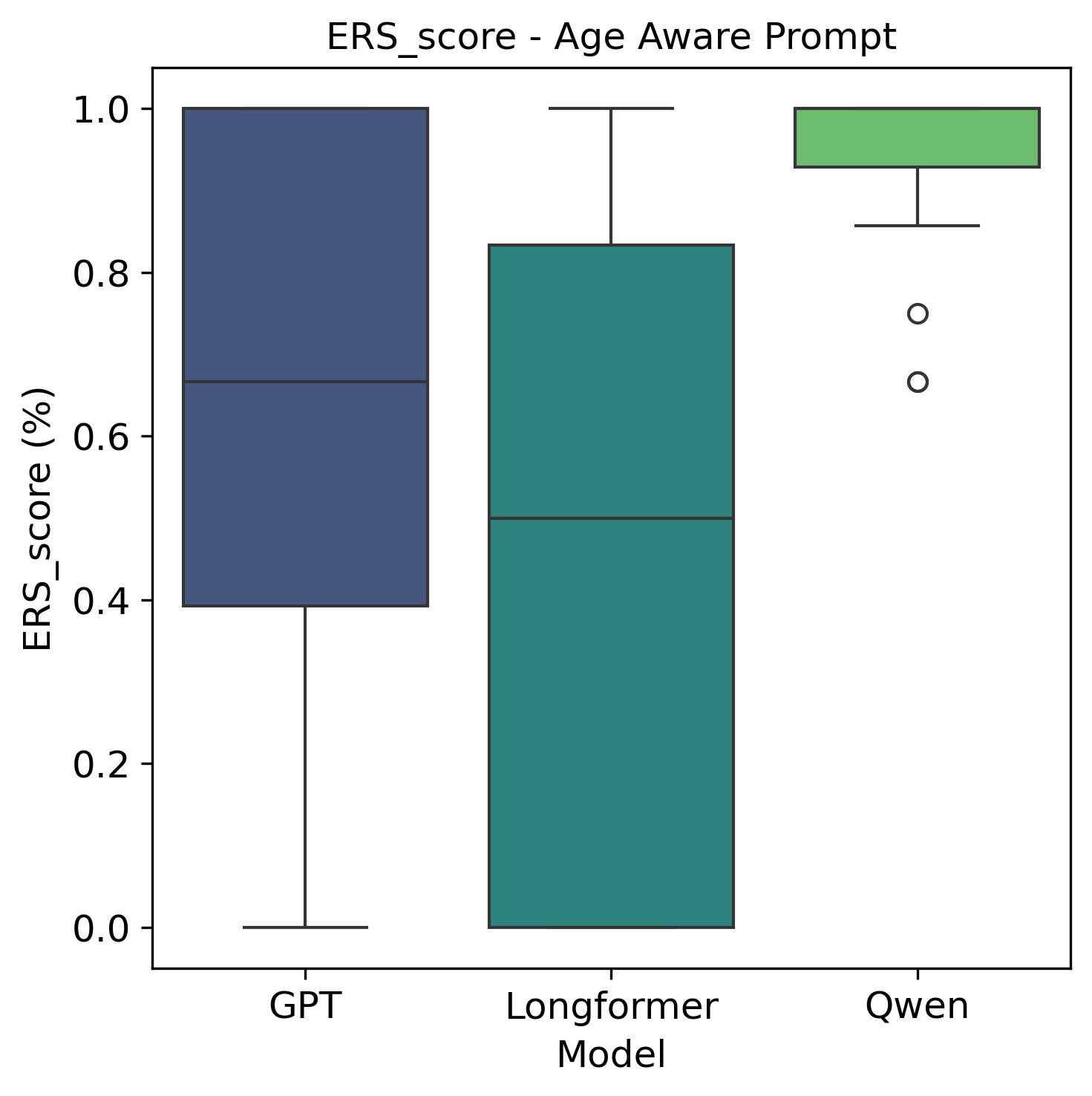}  
    \caption{Entity Retention Across Models (Adult Group) - Age Aware Prompt}
    \label{ers aa}  
\end{figure}

\begin{figure}[!htbp]  
    \centering
    \includegraphics[width=0.9\columnwidth]{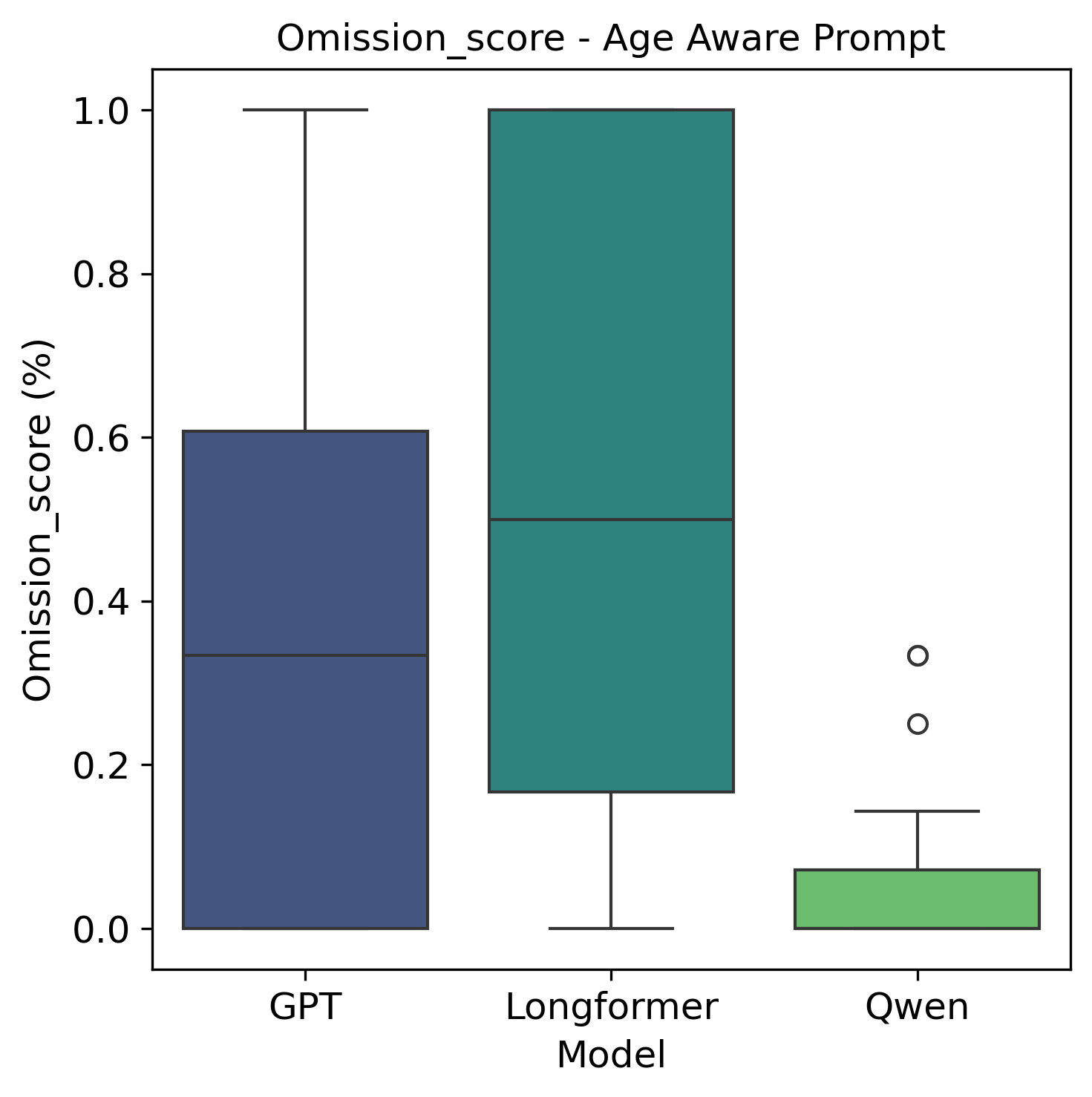}  
    \caption{Entity Omissions Across Models (Adult Group) - Age Aware Prompt}
    \label{omission aa}  
\end{figure}

\begin{figure}[!htbp]  
    \centering
    \includegraphics[width=0.9\columnwidth]{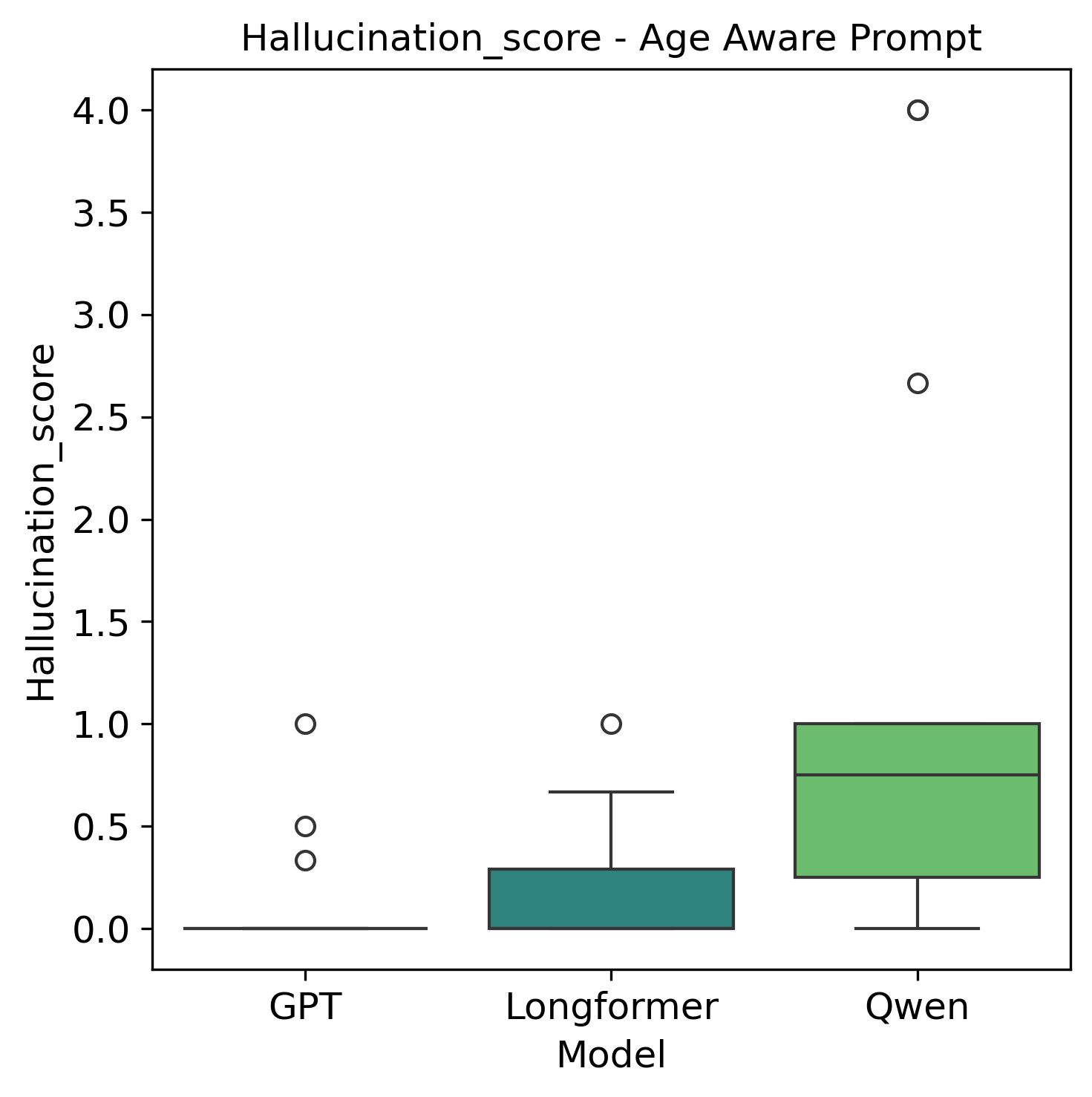}  
    \caption{Hallucinations Across Models (Adult Group) - Age Aware Prompt}
    \label{hal aa}  
\end{figure}

\begin{figure}[!htbp]  
    \centering
    \includegraphics[width=0.9\columnwidth]{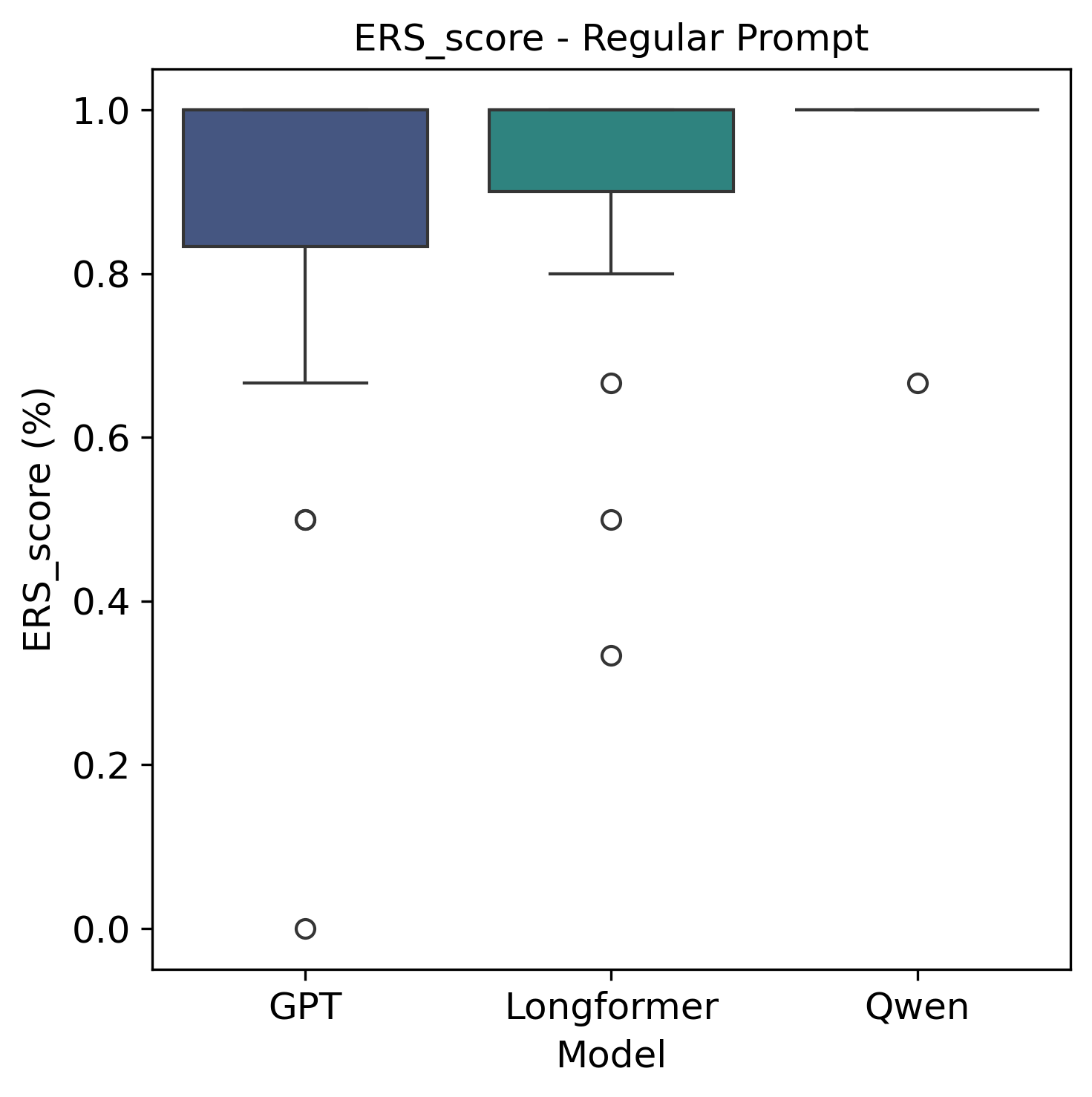}  
    \caption{Entity Retention Across Models (Child Group) - Regular Prompt}
    \label{ers child reg}  
\end{figure}

\begin{figure}[!htbp]  
    \centering
    \includegraphics[width=0.9\columnwidth]{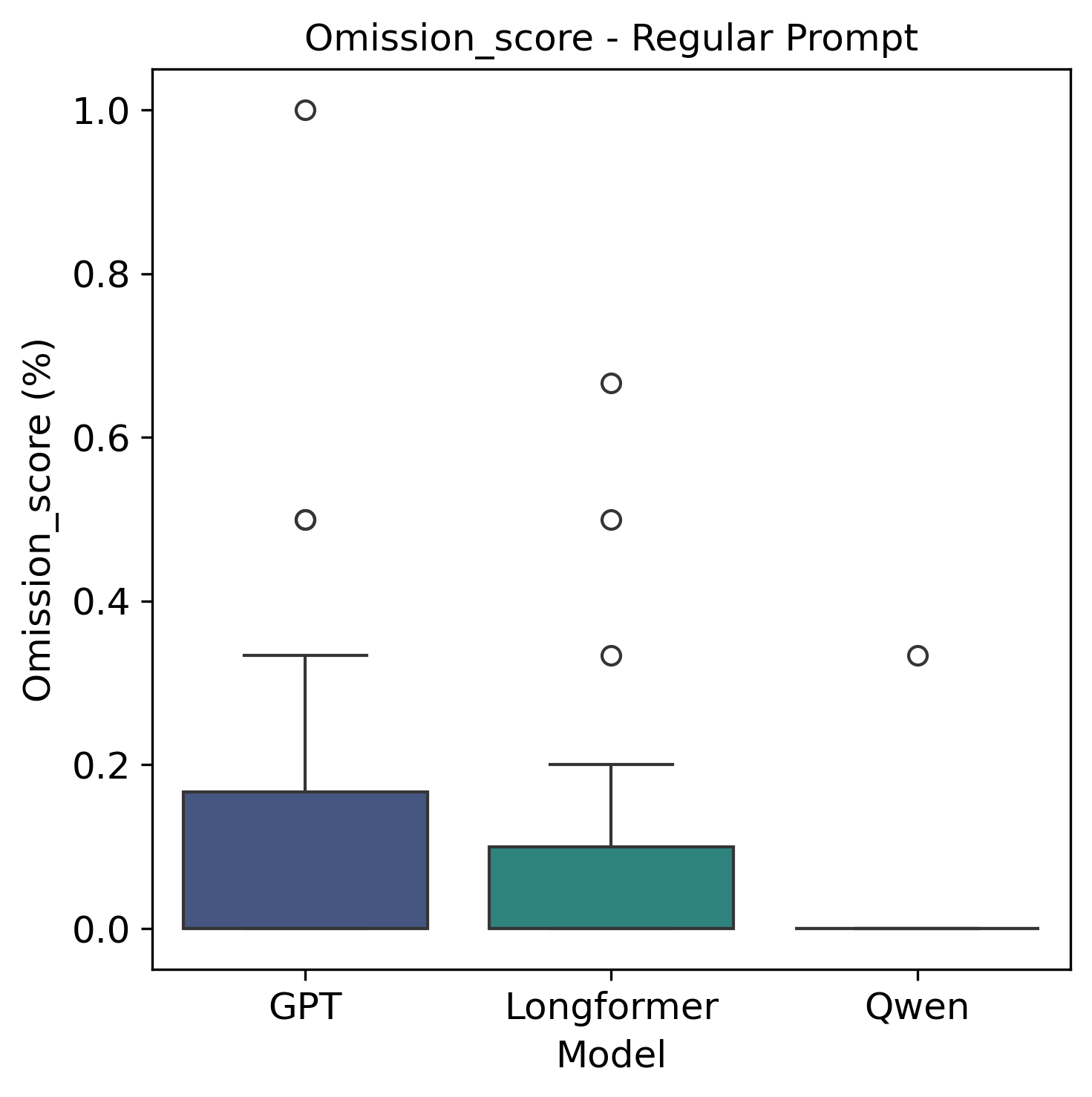}  
    \caption{Entity Omissions Across Models (Child Group) - Regular Prompt}
    \label{omission child reg}  
\end{figure}

\begin{figure}[!htbp]  
    \centering
    \includegraphics[width=0.9\columnwidth]{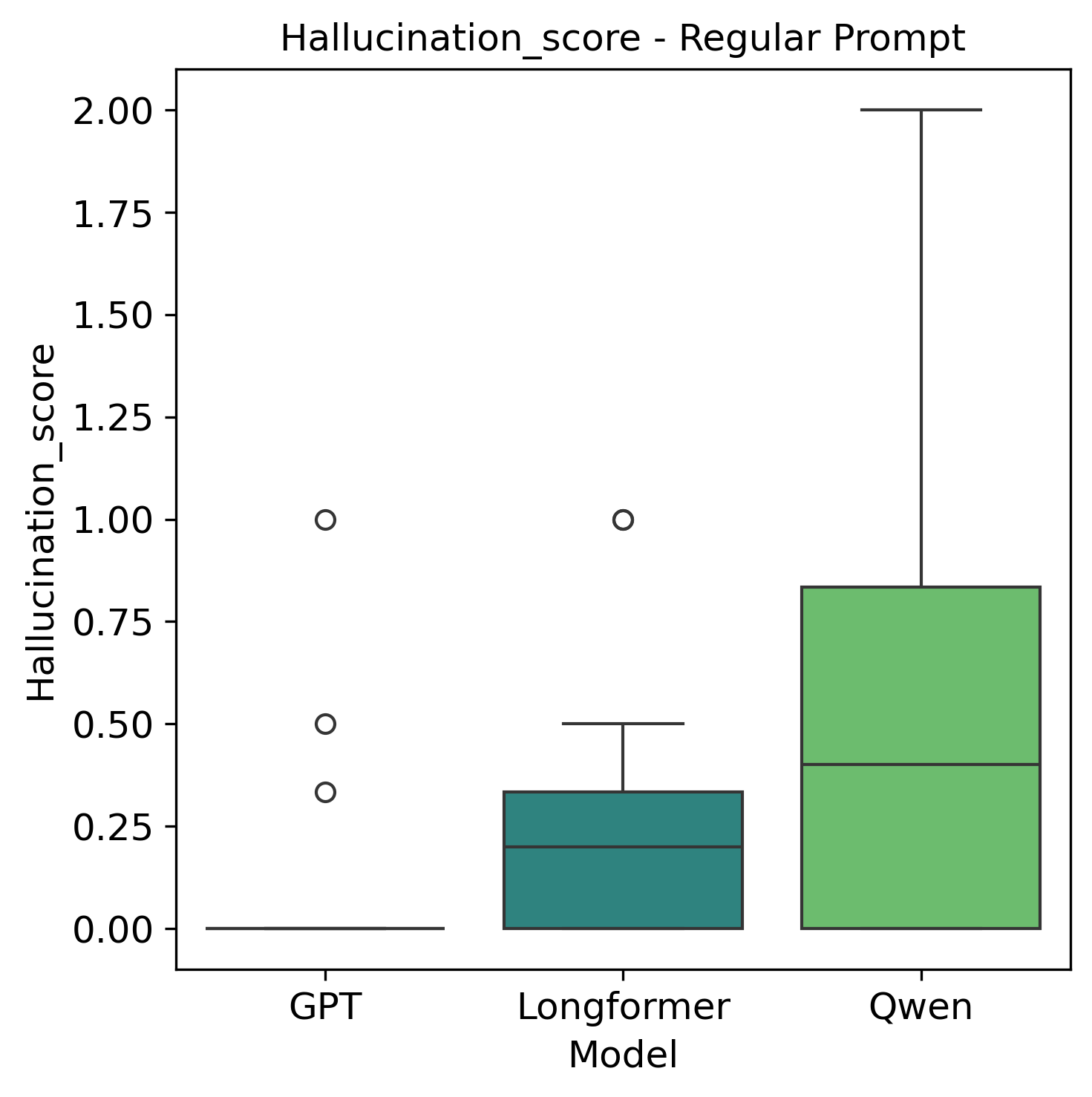}  
    \caption{Hallucinations Across Models (Child Group) - Regular Prompt}
    \label{hal child reg}  
\end{figure}

\begin{figure}[!htbp]  
    \centering
    \includegraphics[width=0.9\columnwidth]{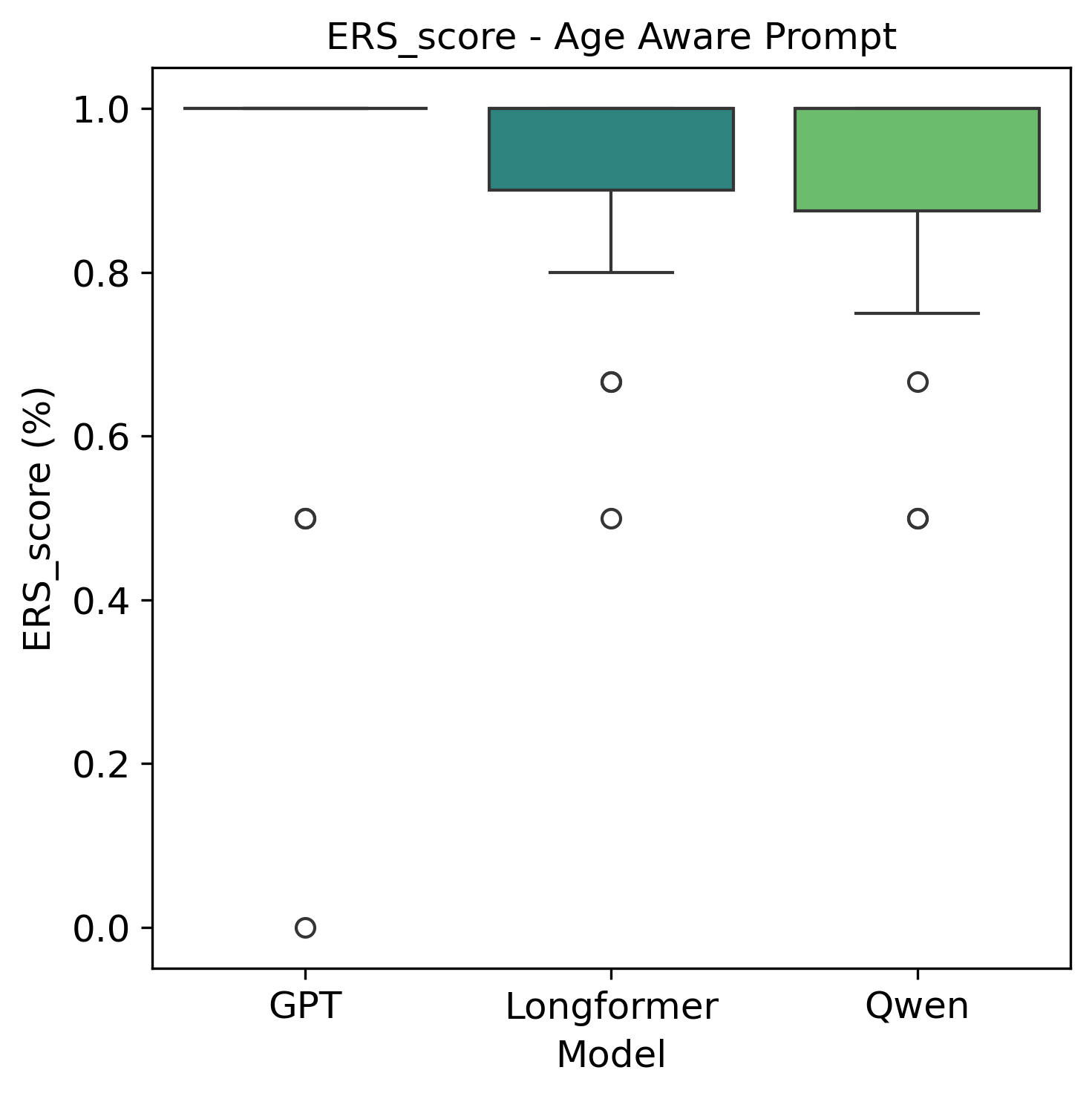}  
    \caption{Entity Retention Across Models (Child Group) - Age Aware Prompt}
    \label{ers child aa}  
\end{figure}

\begin{figure}[!htbp]  
    \centering
    \includegraphics[width=0.9\columnwidth]{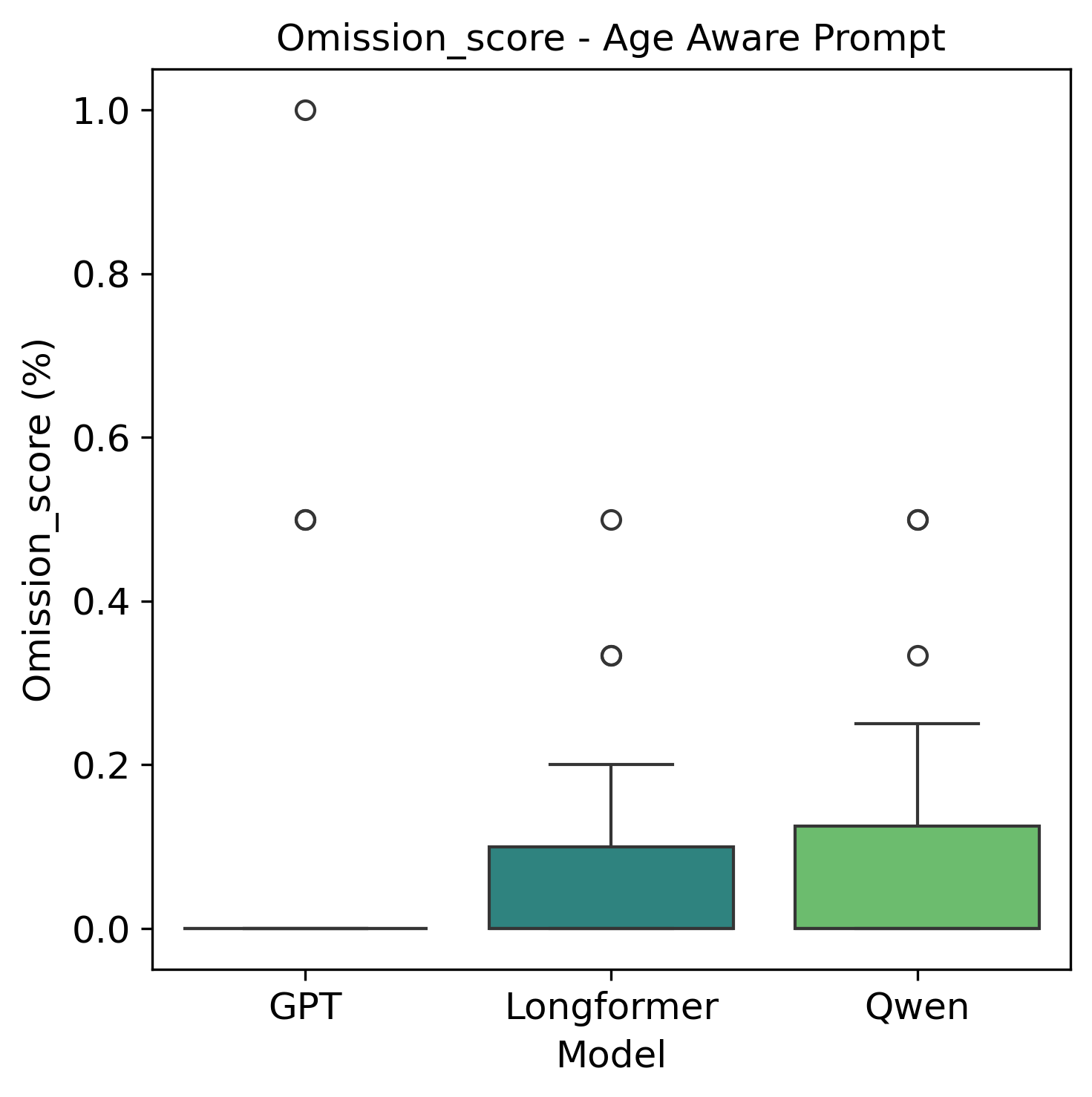}  
    \caption{Entity Omissions Across Models (Child Group) - Age Awaare Prompt}
    \label{omission child aa}  
\end{figure}

\begin{figure}[!htbp]  
    \centering
    \includegraphics[width=0.9\columnwidth]{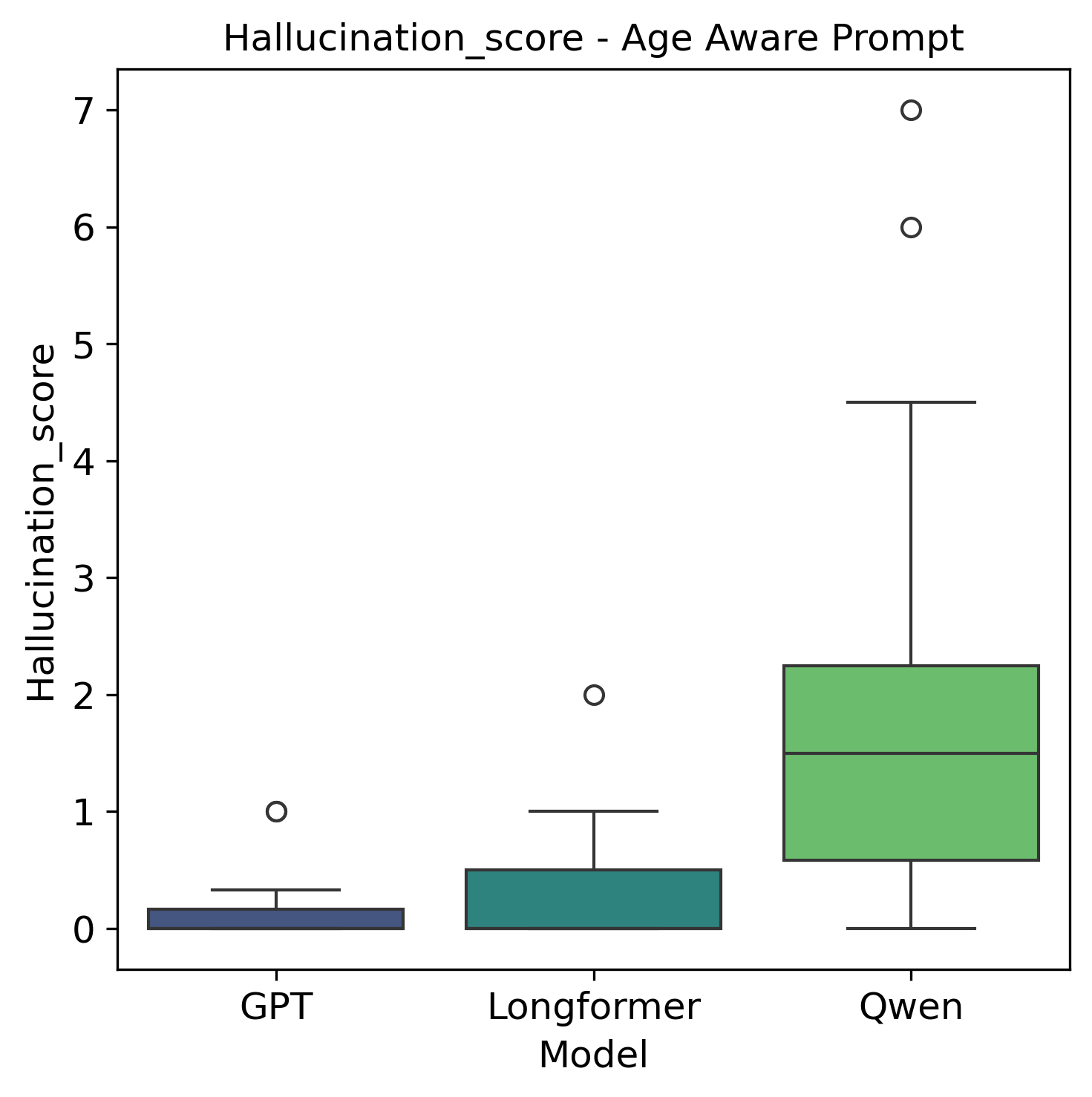}  
    \caption{Hallucinations Across Models (Child Group) - Age Aware Prompt}
    \label{hal child aa}  
\end{figure}

\begin{figure}[!htbp]  
    \centering
    \includegraphics[width=0.9\columnwidth]{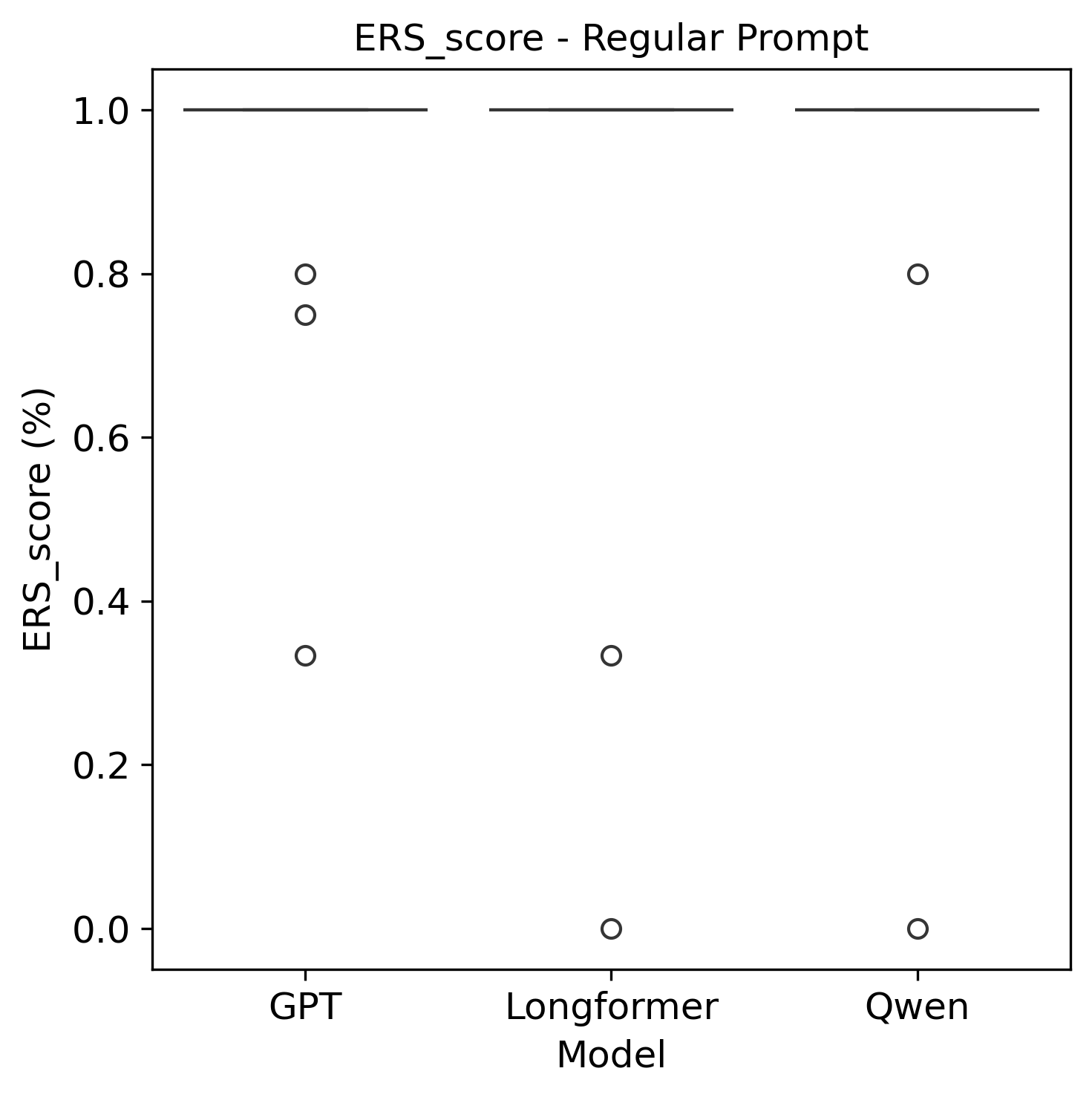}  
    \caption{Entity Retention Across Models (Older Adult Group) - Regular Prompt}
    \label{ers old reg}  
\end{figure}

\begin{figure}[!htbp]  
    \centering
    \includegraphics[width=0.9\columnwidth]{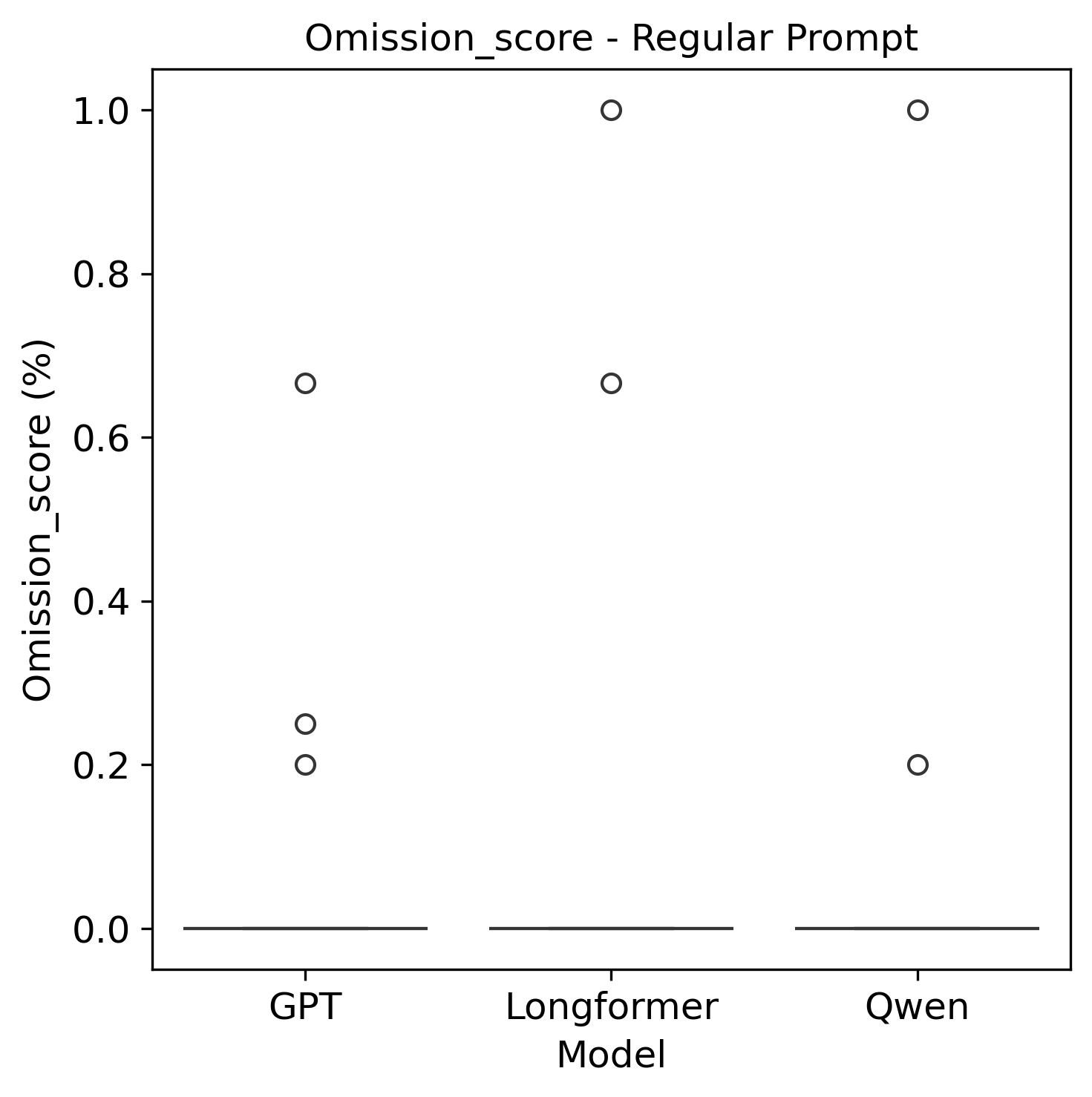}  
    \caption{Entity Omissions Across Models (Older Adult Group) - Regular Prompt}
    \label{omission old reg}  
\end{figure}

\begin{figure}[!htbp]  
    \centering
    \includegraphics[width=0.9\columnwidth]{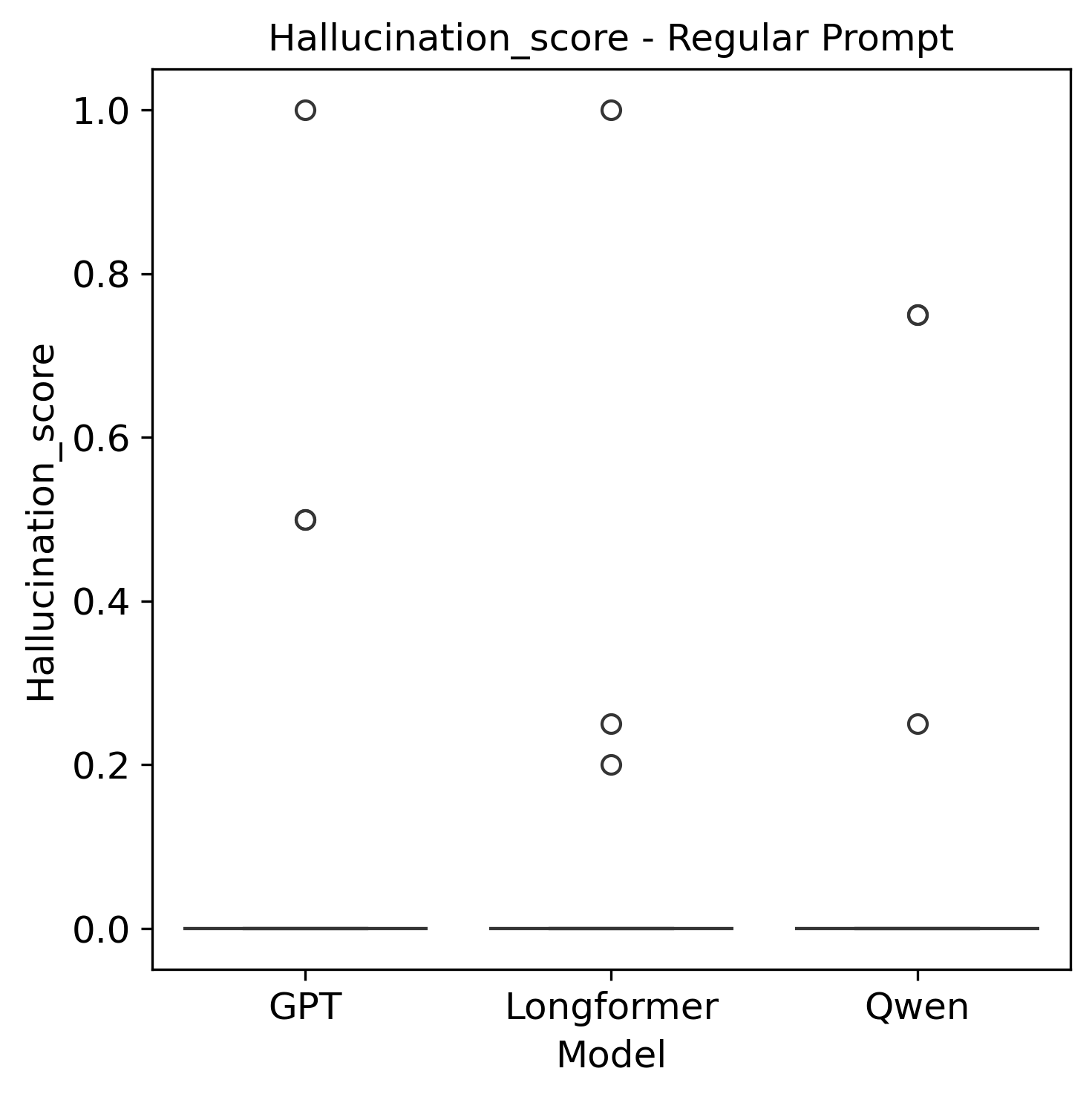}  
    \caption{Hallucinations Across Models (Older Adult Group) - Regular Prompt}
    \label{hal old reg}  
\end{figure}

\begin{figure}[!htbp]  
    \centering
    \includegraphics[width=0.9\columnwidth]{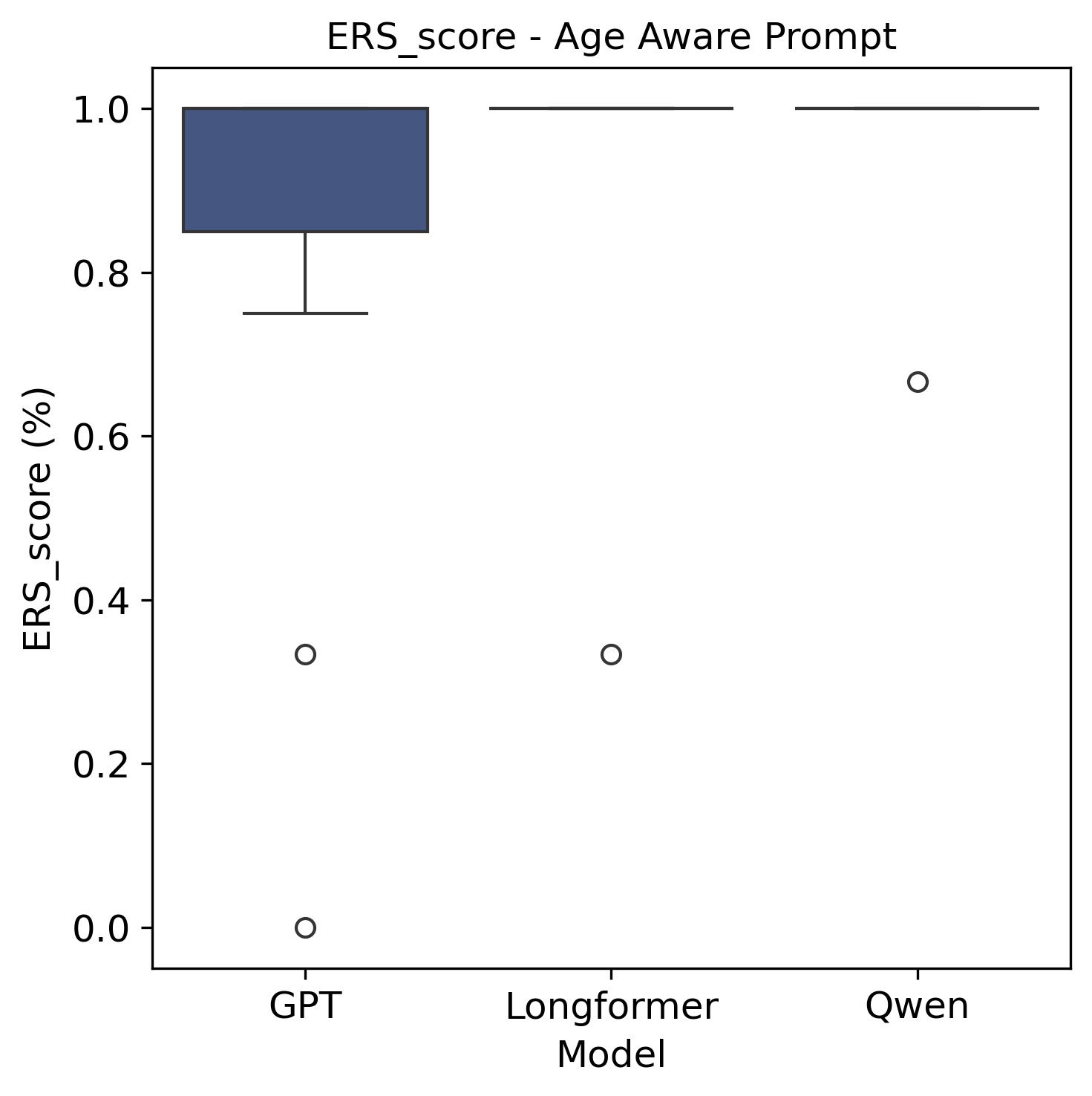}  
    \caption{Entity Retention Across Models (Older Adult Group) - Age Aware Prompt}
    \label{ers old aa}  
\end{figure}

\begin{figure}[!htbp]  
    \centering
    \includegraphics[width=0.9\columnwidth]{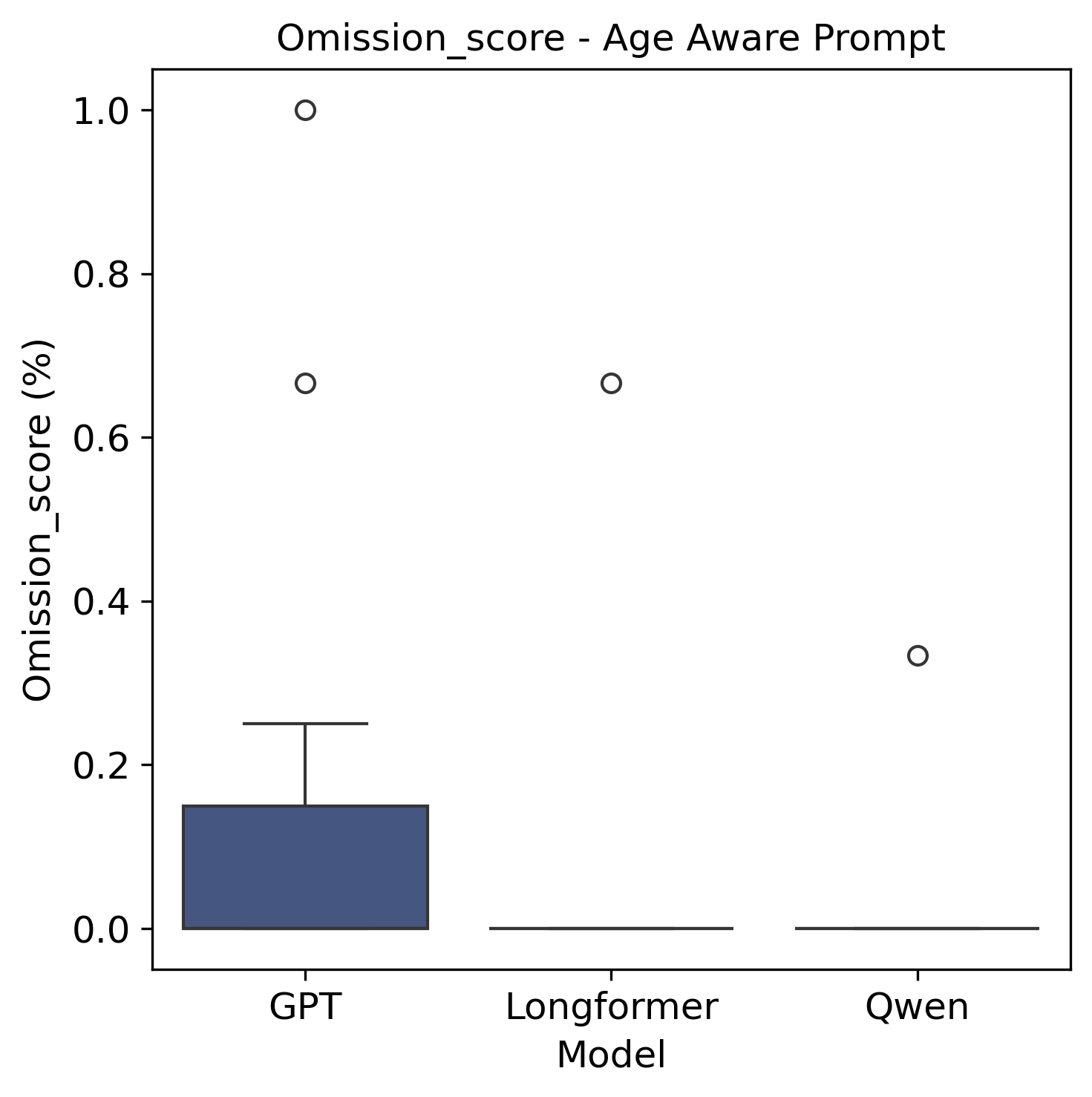}  
    \caption{Entity Omissions Across Models (Older Adult Group) - Age Aware Prompt}
    \label{omission old aa}  
\end{figure}

\begin{figure}[!htbp]  
    \centering
    \includegraphics[width=0.9\columnwidth]{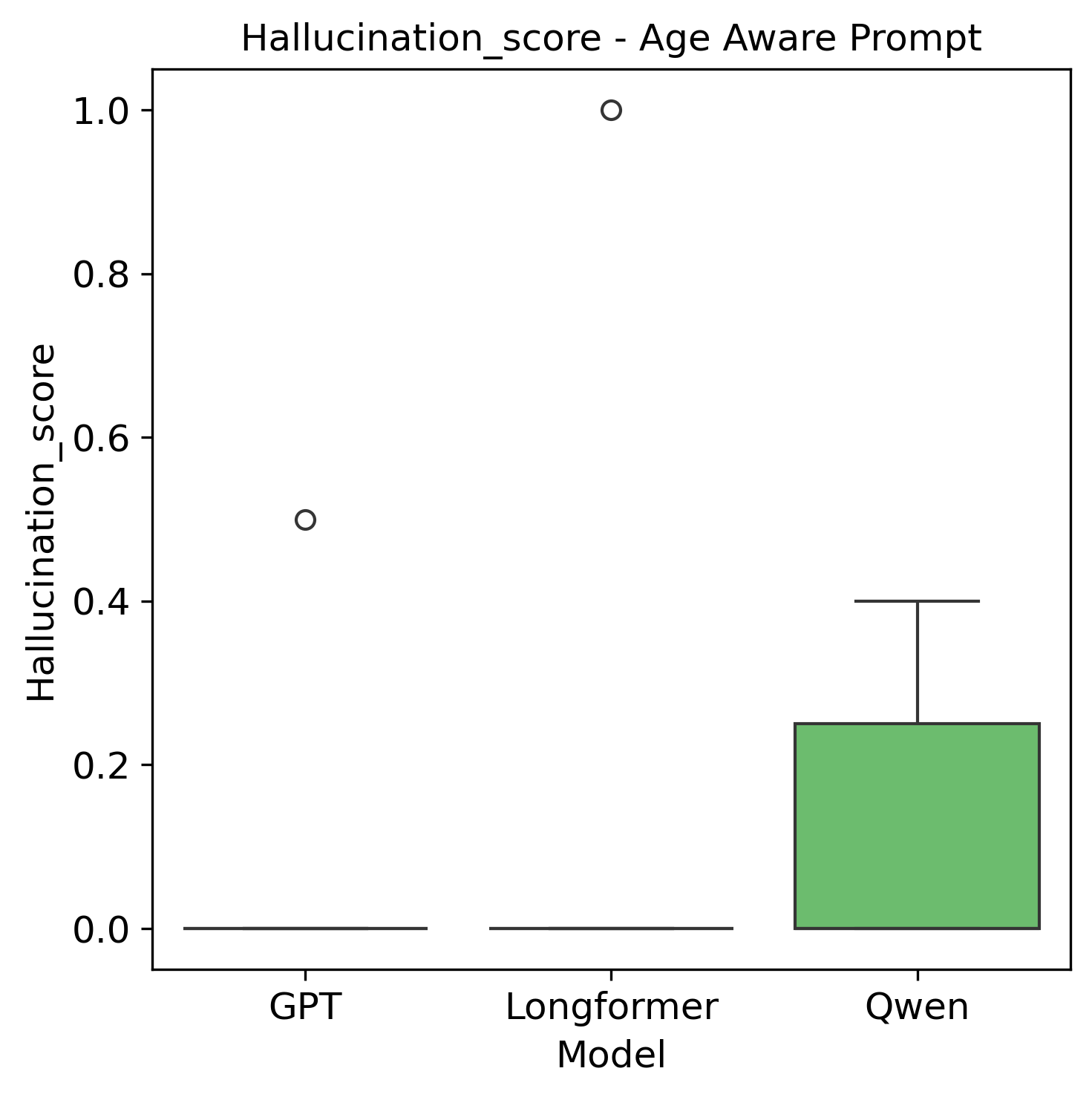}  
    \caption{Hallucinations Across Models (Older Adult Group) - Age Aware Prompt}
    \label{hal old aa}  
\end{figure}
\end{document}